\documentclass[runningheads]{llncs}
\usepackage{graphicx}

\usepackage{amsmath,amssymb} 

\usepackage{color}
\usepackage{comment}
\usepackage{epsfig}
\usepackage{graphicx}
\usepackage{stmaryrd}

\usepackage{booktabs}

\usepackage{makecell}
\usepackage{multirow}
\usepackage{array}
\usepackage{gensymb}

\usepackage{caption}
\usepackage{subcaption}
\usepackage{units}

\usepackage{soul}

\begin{document}

\pagestyle{headings}
\mainmatter

\def\ACCV20SubNumber{632}  

\title{Best Buddies Registration for Point Clouds}

\titlerunning{Best Buddy Registration}

\author{Amnon Drory \and
Tal Shomer \and Shai Avidan \and
Raja Giryes}

\authorrunning{Drory, A. et al.}
%
\institute{Tel Aviv University, Israel}

\maketitle

\begin{abstract}
We propose new, and robust, loss functions for the point cloud registration problem. Our loss functions are inspired by the Best Buddies Similarity (BBS) measure that counts the number of mutual nearest neighbors between two point sets. This measure has been shown to be robust to outliers and missing data in the case of template matching for images. We present several algorithms, collectively named \emph{Best Buddy Registration (BBR)}, where each algorithm consists of optimizing one of these loss functions with Adam gradient descent. The loss functions differ in several ways, including the distance function used (point-to-point vs. point-to-plane), and how the BBS measure is combined with the actual distances between pairs of points. Experiments on various data sets, both synthetic and real, demonstrate the effectiveness of the BBR algorithms, showing that they are quite robust to noise, outliers, and distractors, and cope well with extremely sparse point clouds. One variant, BBR-F, achieves state-of-the-art accuracy in the registration of automotive lidar scans taken up to several seconds apart, from the KITTI and Apollo-Southbay datasets.

\end{abstract}

\section{Introduction}

\begin{figure}
    \centering
    \includegraphics[width=1.5cm]{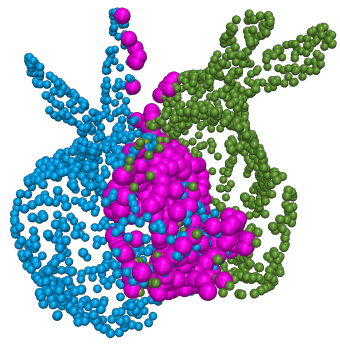}
    \includegraphics[width=1.5cm]{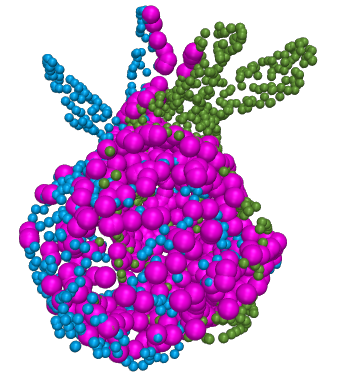}
    \includegraphics[width=1.5cm]{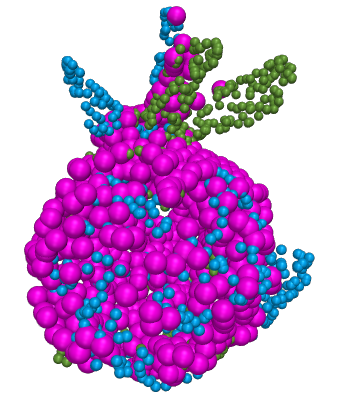}
    \includegraphics[width=1.5cm]{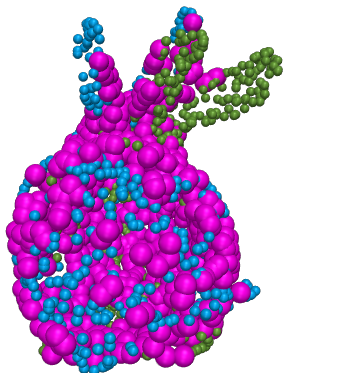}
    \includegraphics[width=1.5cm]{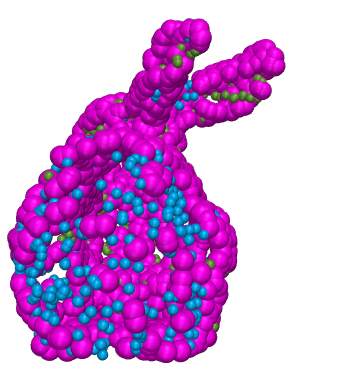}
     \caption{{\bf Best-Buddies Pairs:} The BBR methods perform point cloud registration by iteratively optimizing (with gradient descent) a loss defined by a set of soft or hard \emph{best buddy pairs}. Blue and Green points denote the original point clouds to register. Best buddy pairs are marked in purple. From left to right: iteration 1, 2, 4, 80 and the last iteration (120). } 
     \label{fig:best_buddy_pairs}
\end{figure}

Point clouds registration is an important task in 3D computer vision. The same object or scene is scanned from two view points, e.g. with a laser scanner, and the goal is to recover the rigid transformation (rotation and translation) that aligns the two scans to each other. Realistic scenarios add complications: measurement noise, occlusions due to the change in view point, and outliers due to independent motions of free moving objects in the scene (\emph{distractors}). This makes robustness a central issue for point cloud registration algorithms. 

Probably the most popular approach to solve the problem is using some variant of the Iterative Closest Point (ICP) algorithm \cite{ICP}. This method works by iterating between two stages: first match pairs of points between the two clouds, and then apply the transformation that minimizes a loss defined by the distance between the two points in each pair. The simplest version of ICP uses the Euclidean distance between points, but later versions make use of more complex distance measures to achieve faster and more accurate convergence. 
Some of the most popular and successful variants use local normals to define point-to-plane distance measures. ICP-like methods are typically sensitive to noise, requiring the use of steps such as explicit outlier removal to improve their robustness.  

Recently, Oron {\em et al.} introduced the Best-Buddies Similarity (BBS) measure~\cite{BBS}. BBS counts the number of mutual nearest-neighbors between two point sets. This simple measure was used for template matching between images and proved to be resilient to outliers and occluders. This success motivated us to study how the BBS measure could be adapted to the task of point cloud registration. We suggest several differentiable loss functions inspired by BBS. Our registration algorithms consist of optimizing over these losses with a variant of gradient descent (Adam~\cite{articleAdam}), to recover the parameters of the aligning transformation. We collectively name the resulting algorithms \emph{Best Buddy Registration (BBR)}, and demonstrate their high level of robustness to noise, occlusions and distractors, as well as an ability to cope with extremely sparse point clouds. 
Some of the algorithms are able to achieve very high accuracy in noisy settings where robustness is essential. 

\begin{figure*}
    \centering
	\includegraphics[width=10cm]{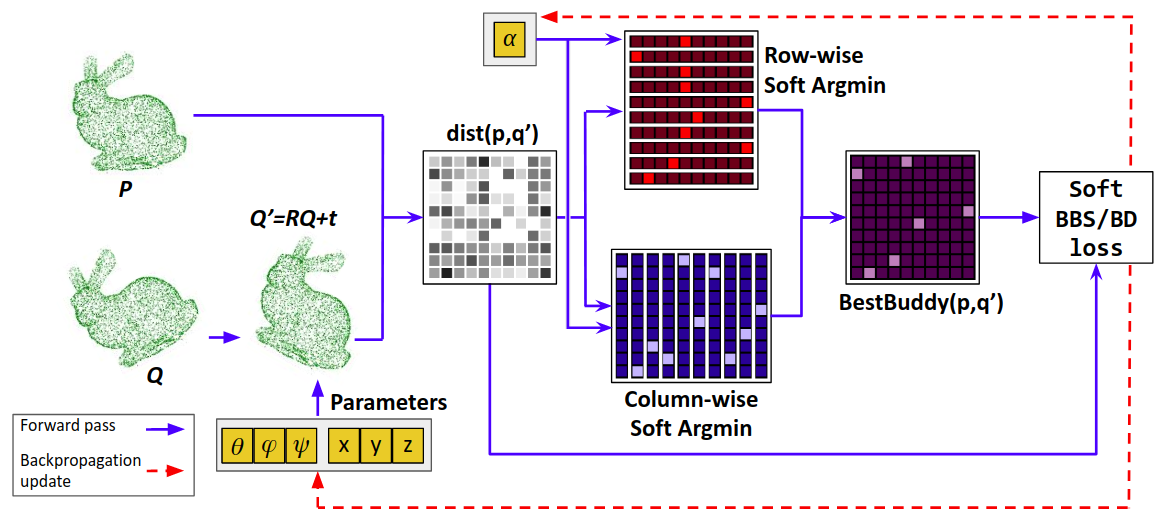}
	\caption{{\textbf Implementation of  Best-Buddies Registration (BBR) as a Neural Network.} Registration of a pair of point clouds is equivalent to "training" this neural network. Unlike in a typical neural network setting, the weights are not learned from a training set. Instead, performing "training" (optimization) on a pair of input point clouds $P$ and $Q$ is equivalent to a gradient-descent search for the optimal rigid transformation between them (equation~1 in main paper). Each forward pass calculates the loss for the current value of $R=R(\theta,\phi,\psi)$ and $t=(x,y,z)$. The back propagation step updates the parameters to improve the match between $P$ and $Q$. The network's weights hold the result of the optimization: the 6 parameters of the transformation, and the temperature parameter $\alpha$ of the soft-argmin function. }
	\label{fig:network_diagram}
\end{figure*}

Deep neural networks (DNN) have increasingly been used for the processing of point clouds lately. BBR can easily be integrated into such DNNs as a registration stage, and be optimized as part of the overall gradient descent optimization of the network.\footnote{For example, the DeepMapping network~\cite{Ding_2019_CVPR} includes a registration stage based on the non-robust Chamfer distance, which could be replaced by BBR.} To facilitate this, we implemented BBR in Pytorch\footnote{https://github.com/AmnonDrory/BestBuddiesRegistration}, which also makes it possible to run the algorithms on widely available neural network infrastructure, such as GPUs. (See figure~\ref{fig:network_diagram}). 

The main contributions of this paper are:
\begin{enumerate}
\item A robust and accurate point cloud registration algorithm that is especially useful in realistic scenarios with a large time offset between the pair of point clouds, meaning large occlusions and outlier motions.
\item The algorithm naturally fits into the deep learning settings as a component: it can be implemented using operations that already exist in deep learning frameworks, and optimized using Adam gradient descent, which is commonly used for neural network optimization
\end{enumerate}

\section{Related Work}

There are various approaches to the problem of point cloud registration. These algorithms can be divided into classic (i.e., non-deep) and deep methods. 

\noindent {\bf Classic methods.} ICP was introduced by Besl and Mckay \cite{ICP}, and Chen and Medioni \cite{Chen:1992}. See the survey of Rusinkiewicz and Levoy~\cite{RUS:2001} or the recent review of the topic  by Pomerelo \emph{et al.} \cite{Pomerleau:2015}. The basic ICP algorithm deals with point-to-point registration, but already~\cite{Chen:1992} considered point-to-plane registration to improve accuracy. This, however, requires the use of normals as an additional source of information.

Segal {\em et al.}~\cite{SegalHT09} later extended ICP to a full plane-to-plane formulation and gave it a probabilistic interpretation. 
Jian and Vemuri \cite{Jian:2011:RPS} proposed a robust point set registration. Their approach reformulated ICP as the problem of aligning two Gaussian mixtures such that a statistical discrepancy measure between the two corresponding mixtures is minimized. It was recently accelerated by Eckart \emph{et al.} \cite{Eckart_2018_ECCV} who introduced a Hierarchical multi-scale Gaussian Mixture Representation (HGMR) of the point clouds. Similarly, 
FilterReg~\cite{FilterReg} is a probabilistic point-set registration method that is considerably faster than alternative methods due to its computationally-efficient probabilistic model. Their key idea is to treat registration as a maximum likelihood estimation, which can be solved using the EM algorithm. With a simple augmentation, they formulate the E step as a filtering problem and solve it using advances in efficient Gaussian filters.

ICP is prone to errors due to outliers and missing data. Thus, a variety of heuristics, as well as more principled methods, were introduced to deal with it. Chetverikov \emph{et al.} \cite{TrimmedICP} proposed a robust version of ICP, termed Trimmed ICP, which is based on Least Trimmed Squares that is designed to robustify the minimization of the error function. Bouaziz \emph{et al.} \cite{SparseICP} used sparse inducing norms to cope with missing points and outliers.

Rusinkiewicz \cite{Rusinkiewicz:2019:ASO} recently introduced a symmetric objective function for ICP that approximated a locally-second-order surface centered around the corresponding points. The proposed objective function achieved a larger basin of convergence, compared to regular ICP, while providing state-of-the-art accuracy.

Fitzgibbon \cite{Fitzgibbon01c} replaces ICP with a general-purpose nonlinear optimization (the Levenberg-Marquardt algorithm) that minimizes the registration error  directly. His surprising finding is that his technique is comparable in speed to the special-purpose ICP algorithm.

Another line of research gives the correspondence problem a probabilistic interpretation. Instead of assuming a one-to-one correspondence, assignments are assumed to be probabilistic. Similar to us, these methods, described next, use gradient descent to find the optimal registration between two point clouds.

The differentiable approximation we take resembles that taken in SoftAssign \cite{rangarajan1997softassign}. There, they solve the correspondence problem, as an intermediate step, using a permutation matrix. Because that matrix is non-differentiable it is replaced with a Doubly-Stochastic Matrix.

EM-ICP \cite{granger2002multi} treats point matches as hidden variables and suggests a method that corresponds to an ICP with multiple matches weighted by normalized Gaussian weights, giving birth to the EM-ICP acronym of the method.

In KCReg \cite{tsin2004correlation}, the authors take an information theoretic approach to the problem. First, they define a kernel correlation that measures affinity between every pair of points. Then, they use that to measure the compactness of a point set and then show that registering two point sets minimizes the overall compactness. In addition, they show that this is equivalent to minimizing Renyi’s Quadratic Entropy. In fact, the only difference between the gradients of KCReg \cite{tsin2004correlation} and EM-ICP \cite{granger2002multi} is the normalization term.

\noindent {\bf Deep methods.} The introduction of PointNet \cite{qi2017pointnet} for processing unordered point clouds led to the development of PointNet-based registration algorithms.

PointNetLK \cite{PointNetLK} maps the two point clouds to some latent space in which it applies the Lucas-Kanade registration \cite{Lucas:1981}. To do that, they define a supervised learning problem that takes two rotated versions of the same point cloud and produces the rotation between the two. The method is implemented using a Recurrent Neural Network, and avoids the costly step of point correspondence. On the downside, it requires a training phase to learn the embedding space, unlike our work that requires no training at all.

Deep Closest Point \cite{Wang_2019_ICCV} consists of three parts: a point cloud embedding network, an attention-based module combined with a pointer generation layer \cite{Vinyals15Pointer} to approximate combinatorial matching, and a differentiable singular value decomposition (SVD) layer to extract the final rigid transformation. 
PointGMM \cite{Hertz20PointGMM} represents the data via a hierarchical Gaussian mixture and learns to perform registration by transforming shapes to their canonical position.
DeepVCP \cite{Lu_2019_ICCV}, for {\it Virtual Corresponding Points}, trains a network to detect keypoints, match them probabilistically, and recover the registration using them. 

A major drawback of deep learning based registration methods is that they strongly depend on the data that they have been trained on. A registration network that is trained for a given dataset does not necessarily generalize well to other datasets \cite{Sarode2019PCRNetPC}. As we do not have a training step, our approach does not suffer from this problem. 

Our approach builds on the work of Oron \emph{et al.} \cite{BBS} and that of Pl\"{o}tz and Roth \cite{plotz2018neural}. Oron {\em et al.} introduced the concept of the best-buddies similarity measure as a robust method for template matching in images. The idea was to map image patches to points in some high dimensional space and count the number of mutual nearest neighbor matches between the two point sets. This was shown to converge to the $\chi^2$ error measure when the number of points tends to infinity.

Pl\"{o}tz and Roth \cite{plotz2018neural} proposed an approximation scheme to the nearest neighbor problem. Instead of selecting a particular element to be the nearest neighbor to a query point, they use a soft approximation that is governed by a temperature parameter. When the temperature goes to zero, the approximation converges to the deterministic nearest neighbor. Similarly to ICP and its variants, the best-buddies similarity relies on nearest neighbor search, and we use this nearest neighbor approximation in our work.


\section{Method}

We consider two point clouds $P=\{p_i\}_{i=1}^{n}$, $Q=\{q_i\}_{i=1}^{m}$, where $p_i, q_j \in \mathbb{R}^3$. We wish to find the transformation that aligns them, and in this work we assume this is a rigid transformation with 6 degrees of freedom (6DOF). We define several differentiable loss functions inspired by the \emph{Best Buddy Similarity measure (BBS)}. We collectively name our registration algorithms \emph{Best Buddy Registration (BBR)}. For each loss function $L$, the algorithm \emph{BBR-L} works by optimizing over this loss function to find the aligning transformation:
\begin{equation}
\label{eq:optimization}
    \arg \min_{R,t,\alpha} {\cal L}(P, RQ+t),
\end{equation}
where $R$ is a 3D rotation matrix, $t$ a 3D translation vector, and $\alpha$ a temperature parameter (discussed ahead). We parameterize the rotation using Euler angles: $R = R(\theta,\phi,\psi)$.  We next describe the four variants of our algorithm: \emph{BBR-softBBS}, \emph{BBR-softBD}, \emph{BBR-N} and \emph{BBR-F}. 



We start by defining the BBS measure: Let $D \in \mathbb{R}^{n \times m}$ denote the distance matrix between points in $P$ and points in $Q$. A best buddies matrix $B$ determines if a pair of points $p_i$ and $q_j$ are mutual nearest neighbors:
\begin{equation}
\label{eq:bbp}
    B_{ij} = \llbracket i = \arg \min_{i'} D_{{i'}j} \rrbracket \cdot \llbracket j = \arg \min_{j'} D_{i{j'}} \rrbracket,
\end{equation}
where $\llbracket \cdot \rrbracket$ equals $1$ if the term in the brackets is true and zero otherwise.

The Best-Buddies Similarity (BBS) loss ${\cal L}_{BBS}(P,Q)$ is negative the number of best buddies pairs\footnote{in the original definition \cite{BBS}, BBS is normalized by $nm$. We omit that here.}:
\begin{equation}
\label{eq:hardBBS}
    {\cal L}_{BBS}(P,Q) = -\sum_{i,j} B_{ij}.
\end{equation}

The best-buddies similarity measure was shown to be very robust to outliers and missing data~\cite{BBS}, in the context of template matching for images. We bring it to 3D point clouds. ${\cal L}_{BBS}$ is a robust measure for the quality of the matching between two point clouds $P$ and $Q$. However, it cannot be used for gradient-descent optimization, because it uses the non-differentiable argmin operator. To overcome this, we use the soft argmin approximation introduced by \cite{plotz2018neural} for Neural Nearest Neighbors Networks. Specifically, $\bar B$ approximates $B$ as follows:
\begin{equation}
\label{eq:softBBpairs}
    {{\bar B}_{ij} = {\frac{e^{\nicefrac{-D_{ij}}{\alpha}}}{\epsilon+\sum_{j}e^{\nicefrac{-D_{ij}}{\alpha}}}} \cdot {\frac{e^{\nicefrac{-D_{ij}}{\alpha}}}{\epsilon+\sum_{i}e^{\nicefrac{-D_{ij}}{\alpha}}}}}, \,\,\,\,\,\,\,\,\,
\end{equation}
where $\alpha$ is a temperature parameter (see ahead), and $\epsilon$ is a small constant used for numerical stability. The matrix $\bar B$ is the element-wise multiplication of row-wise and column-wise soft argmin of the distance matrix $D$. Observe how it corresponds to the brackets in Equation~\eqref{eq:bbp}. The \emph{softBBS} loss is now given by:
\begin{equation}
\label{eq:softBBS}
    {\cal L}_{softBBS}(P,Q) = - \sum_i \sum_j {\bar B}_{i,j}.
\end{equation}

${\cal L}_{softBBS}$ is not only a differentiable approximation to ${\cal L}_{BBS}$, but also a generalization. While $B_{ij}$ is only non-zero if $p_i$ and $q_j$ are mutual nearest neighbors, ${\bar B}_{ij}$ can also be non-zero when, for example, $p_i$ is $q_j$'s 3rd nearest neighbor, while $q_j$ is $p_i$'s 4th nearest neighbor. The value of the temperature parameter, $\alpha$, controls this behaviour. The smaller it is, the more strict ${\bar B}_{ij}$ becomes, meaning more similar to $B_{ij}$. $\alpha$ is also learned during the optimization (together with the other optimized parameters, $R$ and $t$). However, we find it important to initialize it with a reasonable value. A bad choice of $\alpha$ can result in a flat loss, unsuitable for gradient descent. In all our experiments we set $\alpha_{init} =$ $1\mathrm{e}{-2}$ as it generally provides a smooth approximation to $ {\cal L}_{BBS}$, but with a good slope near the minimum. For numerical stability, we allow it to decrease down to $\alpha_{final} = 1\mathrm{e}{-8}$. 

The next loss we suggest is the \emph{soft buddy distance} loss, or ${\cal L}_{softBD}$, which makes use of the distance between the two points in each pair. This is in contrast to the BBS measure, which only counts the \emph{number} of best-buddies, and softBBS loss, which is a soft approximation to that. We define \emph{softBD} as follows:
\begin{equation}
\label{eq:softBD}
    {\cal L}_{softBD}(P,Q) = \frac{\sum_i \sum_j {\bar B}_{i,j}D_{ij}}{\sum_i \sum_j {\bar B}_{i,j}}.
\end{equation}

The next loss is \emph{softBD with normals} loss, or $\cal{L}_{N}$, which uses a point-to-plane distance measure calculated from local normals. Such a distance measure is used in some of the most popular and successful ICP variants, such as generalized ICP~\cite{SegalHT09}, and symmetric ICP~\cite{Rusinkiewicz:2019:ASO}. In the previous methods we suggested, we used the Euclidean point-to-point distance to create the distance matrix $D$. In \emph{BBR-N} we replace that with the following symmetric point-to-plane distance, based on Rusinkievicz~\cite{Rusinkiewicz:2019:ASO}:
\begin{equation}
\label{eq:normals}
D^{n}_{i,j} = dist(Rq_i+t,p_j) = { |\langle{Rq_i+t-p_j}, {Rn_{q_i}+n_{p_j}}\rangle| }.
\end{equation}
where $n_{q_i}$ is the normal at point $q_i$,  $n_{p_j}$ is the normal at point $p_j$,  and $D^{n}$ denotes the version of the distance matrix calculated using this distance. The loss function is then calculated as in \emph{softBD}, except using $D^{n}$ instead $D$. This distance is symmetric in the sense that it uses the normals from both points, unlike algorithms such as point-to-plane ICP~\cite{Chen:1992} that only use normals from one of the point clouds.\footnote{This is not to be confused with the symmetric nature of the best-buddies similarity measure that we introduce here.}

The final loss we present is the \emph{best-buddy filtering} loss, or ${\cal L}_{F}$. At the heart of the BBS measure lies the robustness achieved by using mutual nearest neighbors. \emph{BBR-F} translates this idea into \emph{best buddy filtering}: using only pairs that are mutual nearest neighbors. In addition, it follows the trend in ICP-like algorithms, in that it uses both point-to-point and point-to-plane distance measures: the Euclidean point-to-point distance is used for the stage of matching pairs between the two point clouds. Then the symmetric point-to-plane distance between these pairs is used to define the following loss:
\begin{equation}
\label{eq:softBBSwithNormals}
    {\cal L}_{F}(P,Q) = - \sum_i \sum_j  B_{i,j} \cdot D^{n}_{i,j}.
\end{equation}
Notice that in this variant of BBR we have a hard selection of pairs, which isn't optimized during gradient descent. Instead, the \emph{distances} between the points in each pair are minimized. \\
The central difference between \emph{BBR-F} and symmetric ICP~\cite{Rusinkiewicz:2019:ASO} is that best buddy filtering replaces explicit outlier rejection. This removes the necessity to calibrate outlier-rejection parameters, while resulting in better accuracy in settings where robustness is important, as Section~\ref{sec:exp} shows. Another difference is that \emph{BBR-F} uses Adam gradient descent for optimization, while symmetric ICP uses a closed form solution to an approximate linearized version of the symmetric point-to-plane distance measure.

\emph{BBR-F} is especially useful for very large point clouds, where memory and running time become a constraint, because it does not require the full distance matrix $D$ or $D^{n}$. For the pair matching step, we use the KD-tree method~\cite{kdtree}, and then calculate the point-to-plane distances only for best-buddy pairs.

\section{Experiments}
\label{sec:exp}

We present experiments that are designed to analyze the behaviour of the BBR methods, and evaluate them on several datasets including Stanford~\cite{Turk:1994:ZPM}, TUM RGBD \cite{DBLP:conf/iros/SturmEEBC12}, KITTI Odometry~\cite{Geiger2012CVPR}, and Apollo-Southbay~\cite{apollo}. We compare our approach to several established alternatives, focusing on classic approaches (e.g., ICP) as opposed to learned methods, as the latter do not necessarily generalize well across datasets \cite{Sarode2019PCRNetPC}.

Figure~\ref{fig:best_buddy_pairs} shows the best-buddies pairs during a typical run of \emph{BBR-softBBS} on the Stanford Bunny model~\cite{Turk:1994:ZPM}. The algorithm converges after 120 iterations. At first, there are few best-buddies pairs, but as time progresses their number grows until convergence. 

\subsection{Performance Evaluation Setup}
\label{subsec:PerformanceEvaluation}

We conduct a set of experiments to evaluate accuracy and robustness. To do that, we apply different rotations, translations, sub-sampling, and noise, to each point cloud. We compare ourselves to the following popular point cloud registration algorithms: (i) HGMR~\cite{Eckart_2018_ECCV}, a GMM based method; (ii) Coherent Point Drift (CPD)~\cite{Myronenko:2010:PSR}, a probabilistic algorithm based on GMM; (iii) Generalized ICP (G-ICP)~\cite{SegalHT09}, a very popular and accurate ICP variant that uses local normals, and (iv) Symmetric ICP  (Sym-ICP)~\cite{Rusinkiewicz:2019:ASO}, which provides state-of-the-art performance on several point cloud registration challenges. 

ICP algorithms are sensitive to noise, and therefore commonly employ a set of standard practices for outlier rejection. Our BBR methods require no such processes. 
\paragraph{\bf{Setting.}} We test the 4 variants of our algorithm: \emph{BBR-softBBS}, \emph{BBR-softBD}, \emph{BBR-N} and \emph{BBR-F}. \emph{BBR-softBBS} tends to be less accurate than the others, and therefore we omit it from most experiments.

The local normals that are used in \emph{BBR-N} and \emph{BBR-F} are estimated by calculating the principal axis of a neighborhood of $k=13$ neighbors around each point in the full cloud (before subsampling). For consistency, Sym-ICP was given the same normals used for \emph{BBR-N}. G-ICP calculates its own normals on-the-fly, from the subsampled point cloud that it takes as input. For all BBR methods, the optimization is performed by running Adam for a pre-defined number of iterations.

We use the Probreg\footnote{https://github.com/neka-nat/probreg} library's implementation of \emph{HGMR} and \emph{CPD}, and the original \emph{SymICP} implementation, for all of which we use the default parameters. We use the Point Cloud Library's~\cite{PCL} implementation of \emph{G-ICP}, setting the parameters as in~\cite{Lu_2019_ICCV}.

In all our experiments, we follow Lu et al.~\cite{Lu_2019_ICCV} in defining angular distance as \emph{chordal distance}  \cite{Hartley2012RotationA}, and translational distance as the Euclidean norm of the difference between two translation vectors.

\subsection{Comparing Accuracy and Robustness between BBR variants}
\label{subsec:BBR_methods}
We start by comparing the different variants of BBR in a simple experiment, testing their accuracy and ability to converge with different initial rotations: 5, 10, 30, 60 and 90 degrees. For each angle, we repeat 20 times: select two random subsets of 500 points from the Stanford Bunny point cloud, rotate the target point cloud around a random axis, perform registration and measure the angular error. 
We consider registration to have failed if the final error is over $5^{\circ}$. 

\begin{figure}[t]
    \includegraphics[width=6cm]{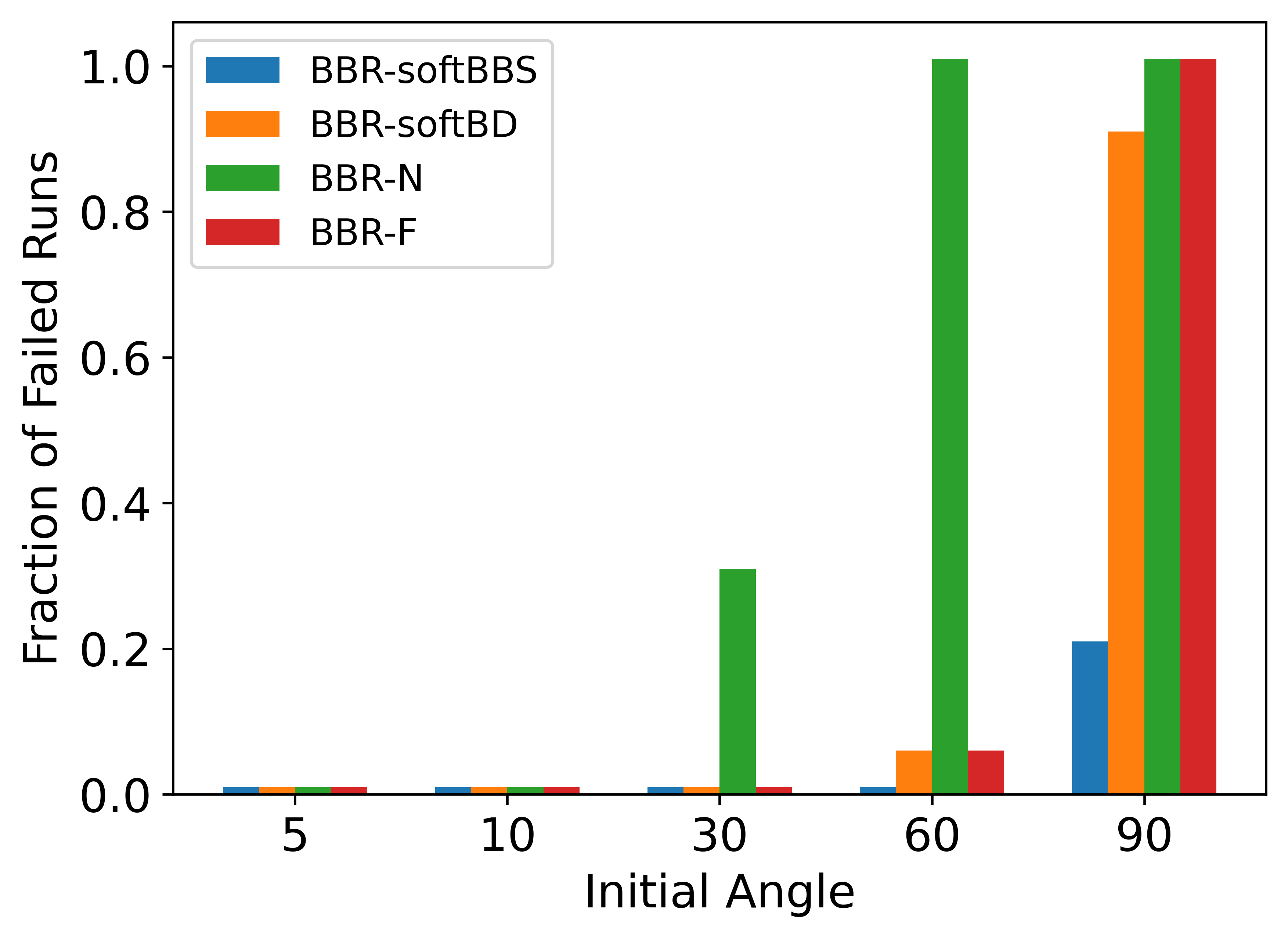}
    \includegraphics[width=6cm]{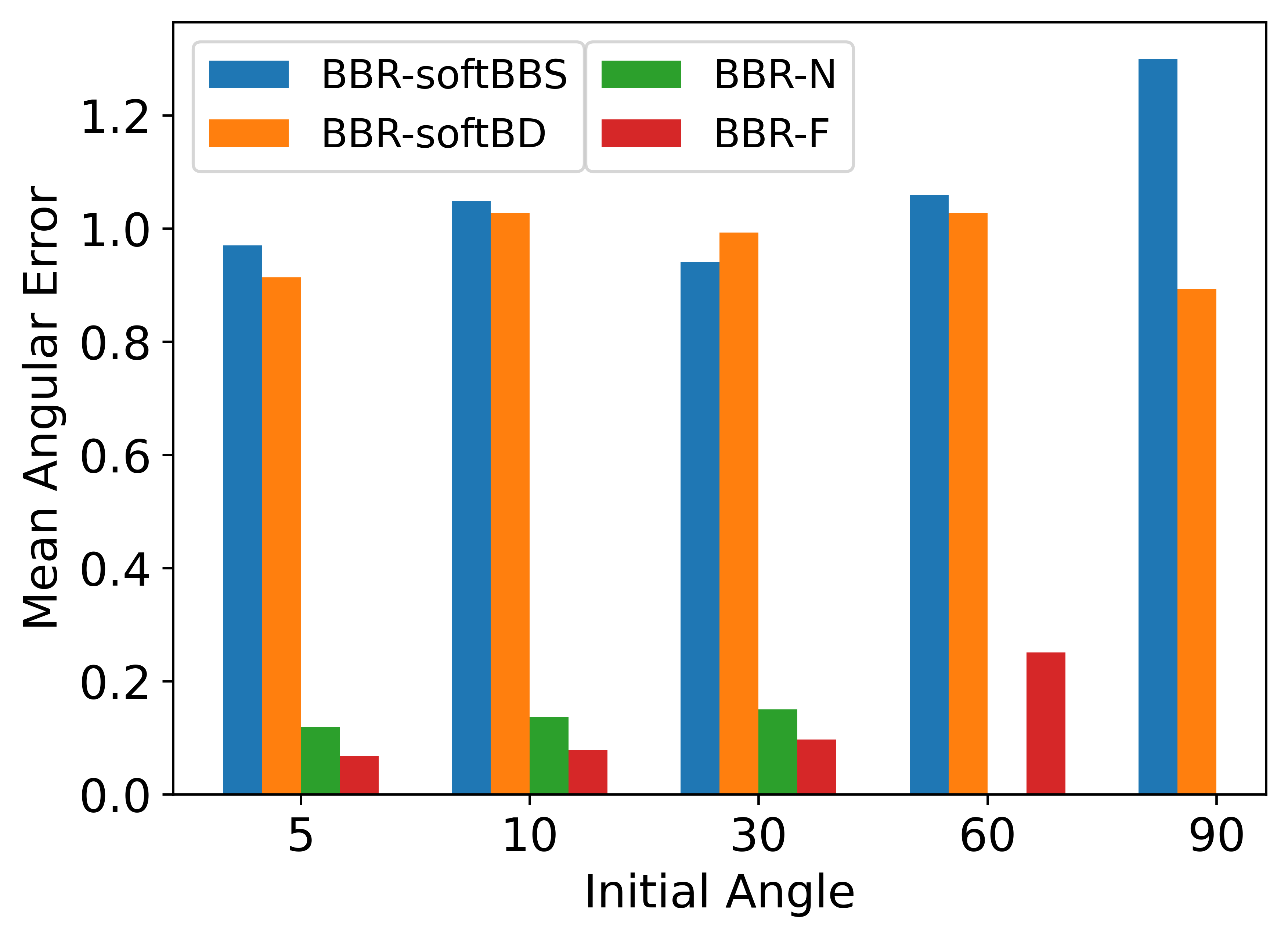}    
    \caption{{{\bf{Accuracy and Robustness to Initial Error}} Comparison among the BBR methods using the Stanford Bunny point cloud. Left - the fraction out of 20 registration attempts that failed, Right - accuracy as measured by the mean angular error of the successful attempts. \emph{BBR-softBBS} is the best at converging from a large initial error. \emph{BBR-N} and \emph{BBR-F}, that make use of local normals, are the most accurate.}}
    \label{fig:exp_initial_angle}
\end{figure}

Results are shown in figure~\ref{fig:exp_initial_angle}. \emph{BBR-softBBS} is the clear leader in robustness to large initial error. Unlike the others, it is able to handle initial rotations of $90^{\circ}$ with hardly any failures. Notice that BBR-N is especially sensitive to large initial rotations, due to its reliance on normals to recognize best-buddies. When the initial rotation is large, the same object will have very different normals in each of the two scans.
For all algorithms, large initial rotations do not cause degradation in accuracy - for the attempts that did succeed. 
When looking at the accuracy of the successful attempts, it's clear that the methods that make use of local normals (\emph{BBR-N} and \emph{BBR-F}) are significantly more accurate. 

\subsection{Accuracy}
\label{subsec:Accuracy}

\begin{figure}[t]
    \centering
    \includegraphics[width=2.5cm]{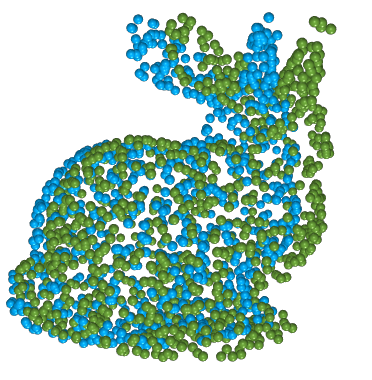}
    \includegraphics[width=3cm]{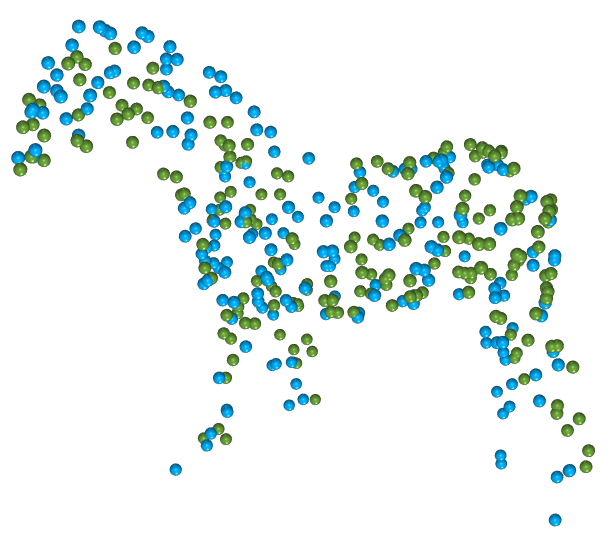}
    \includegraphics[width=3cm]{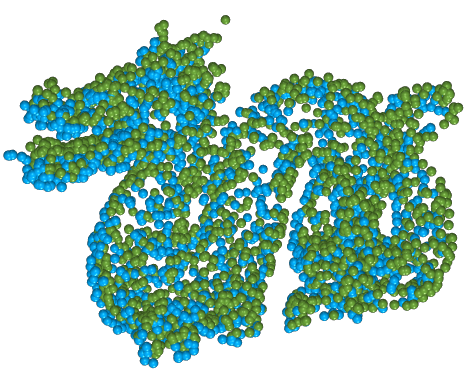}\\
    \includegraphics[width=3.9cm]{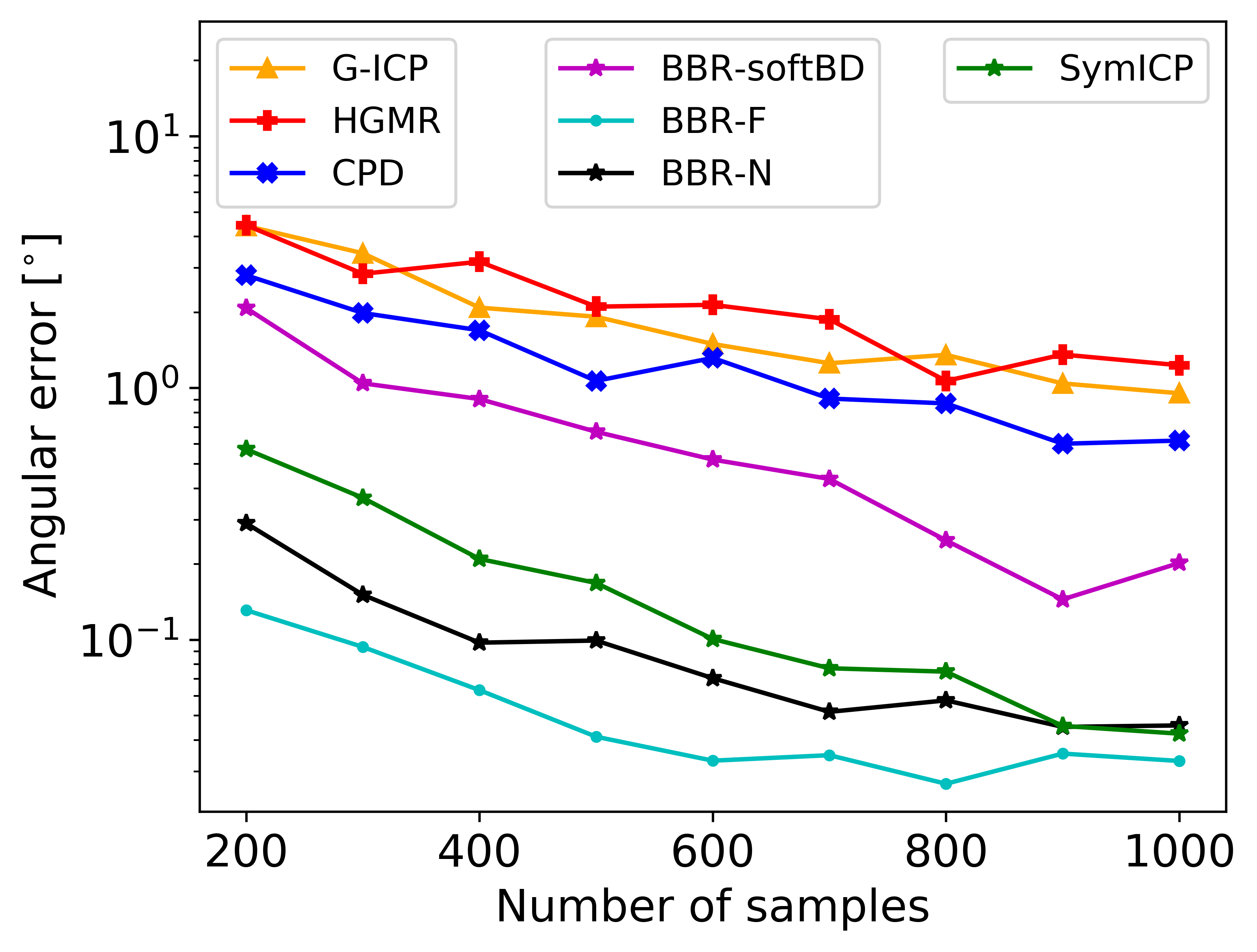} 
    \includegraphics[width=4cm]{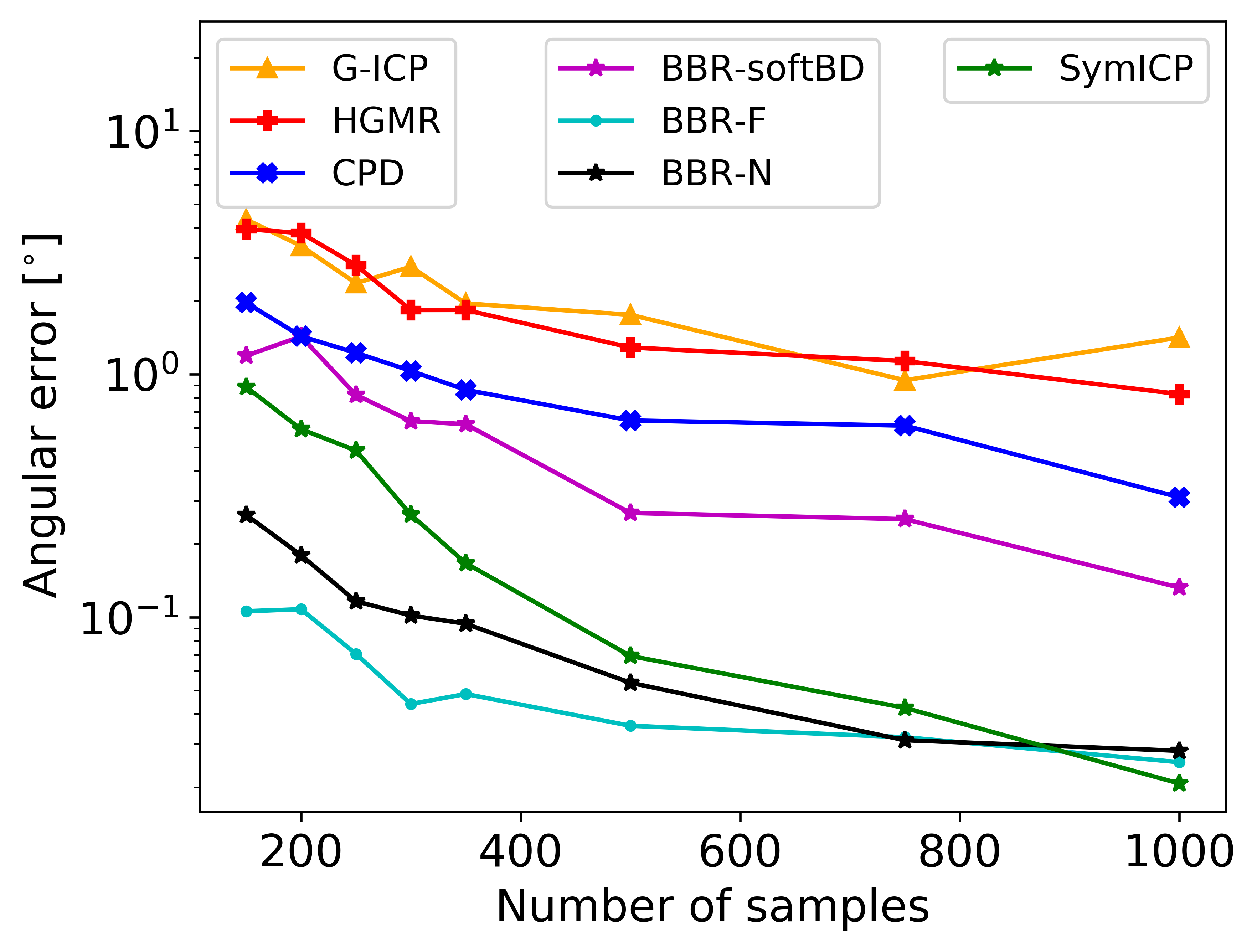}
    \includegraphics[width=4.05cm]{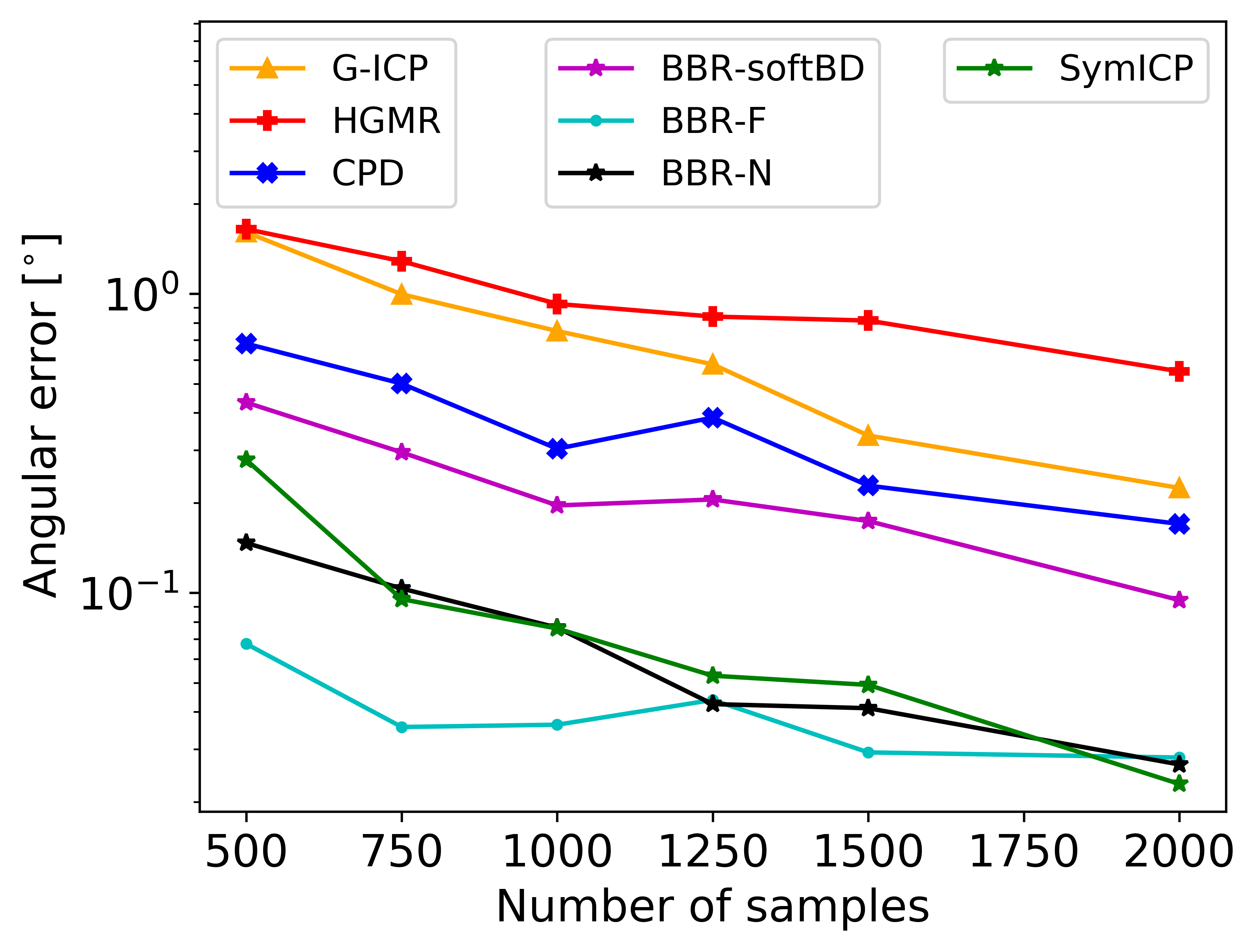} \\
    \includegraphics[width=3.9cm]{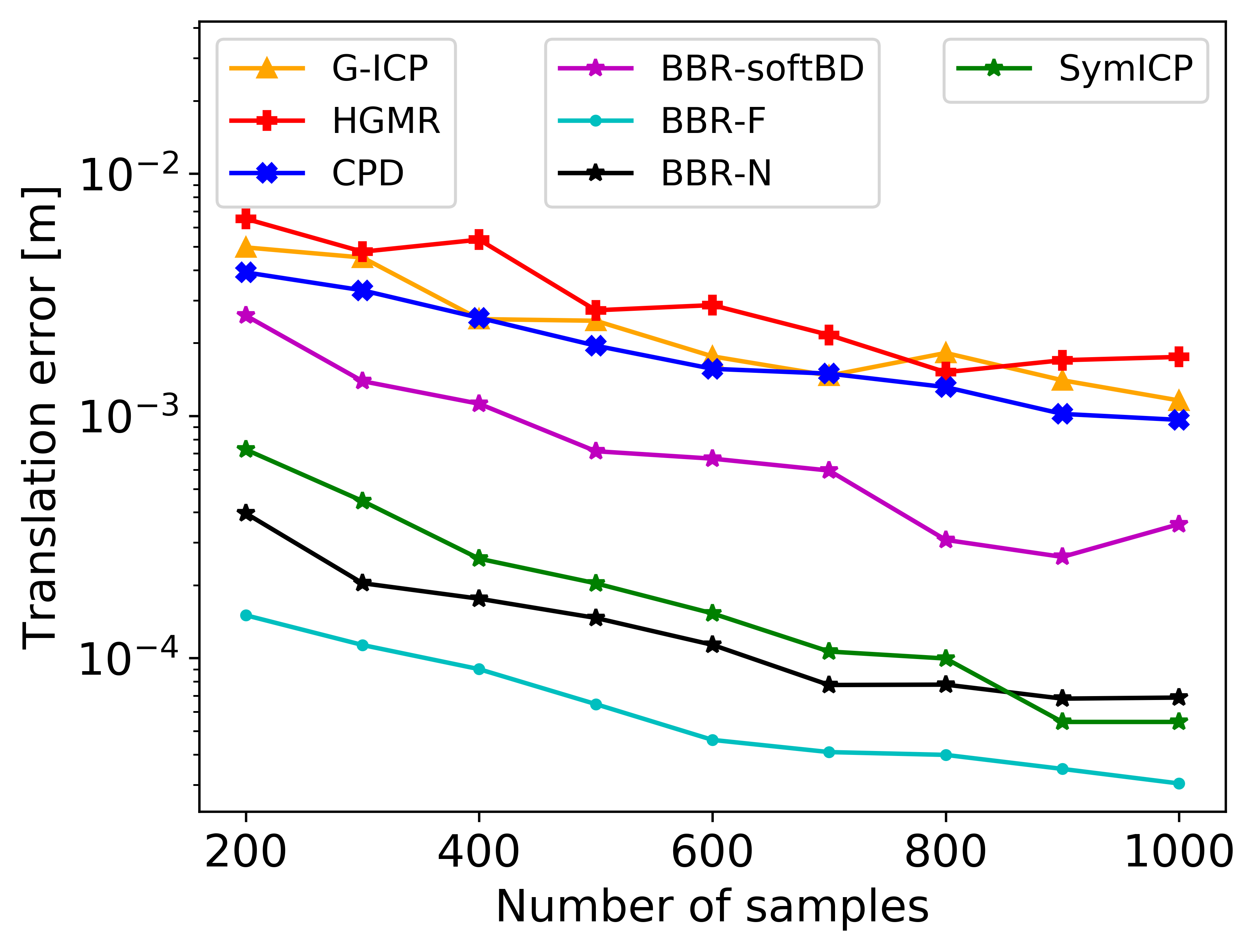} 
    \includegraphics[width=4cm]{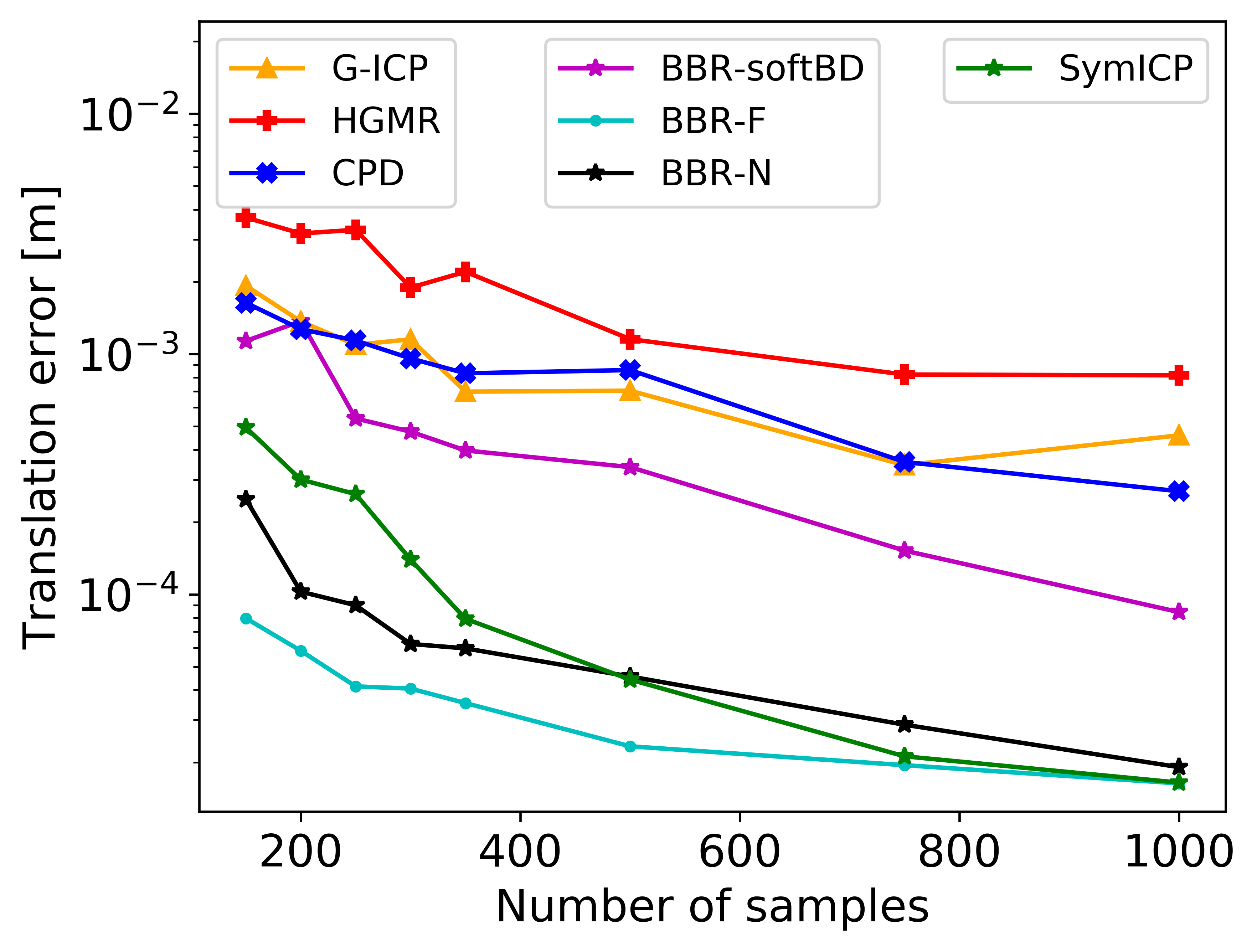}
    \includegraphics[width=4.05cm]{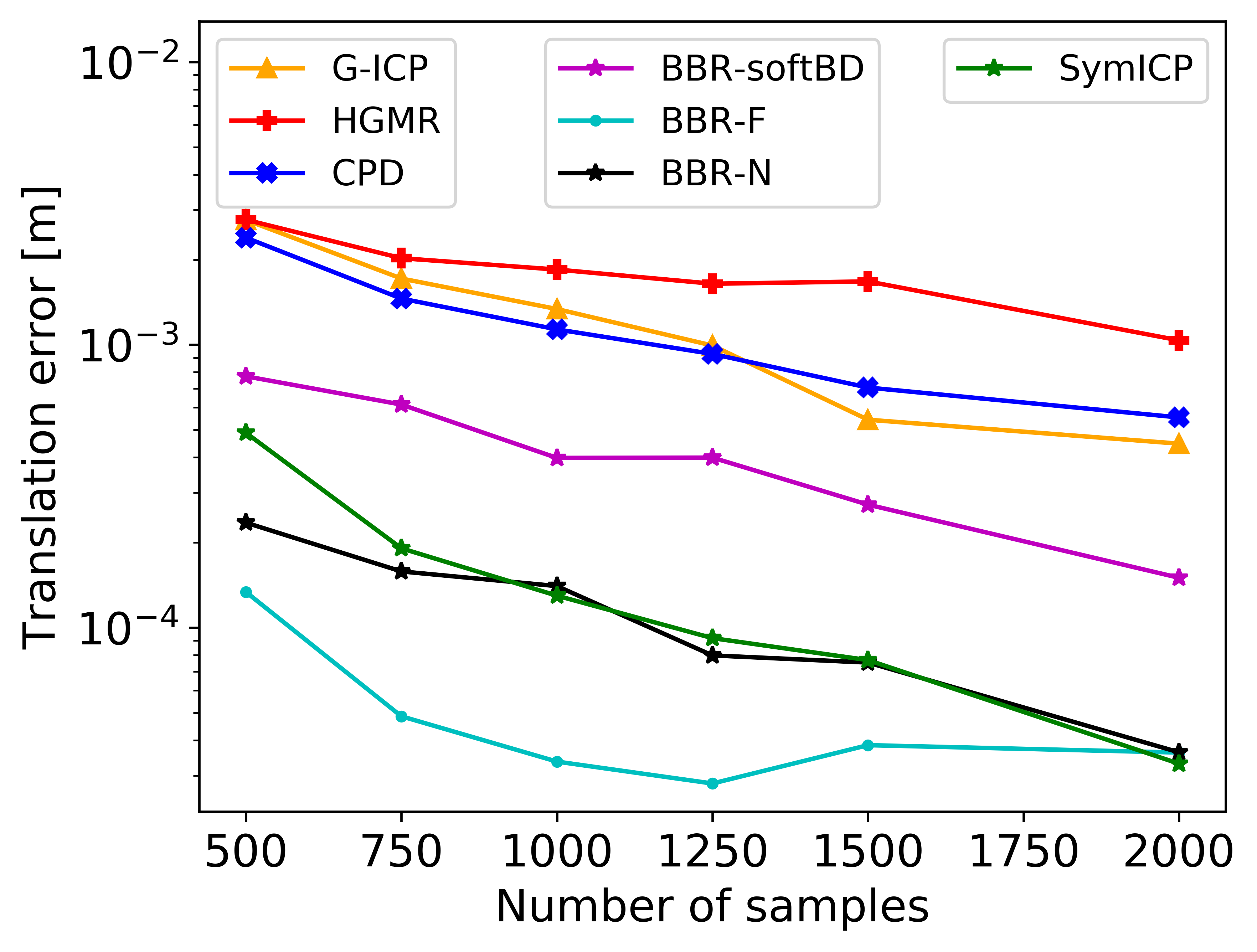}
     \caption{{\bf Accuracy test:} Top Row - Point clouds examples as used in the accuracy test. Middle and bottom rows - angular and translation error as a function of the number of points. \emph{BBR-F} achieves the most accurate results in all scenarios, followed closely by \emph{BBR-N}. \emph{SymICP} becomes competitive only when the number of points increases. } 
    \label{fig:accuracy}
\end{figure}

The experiments shown in Figure~\ref{fig:accuracy} demonstrate our ability to register point clouds with random rotation and translation, using the Bunny, Horse, and Dragon point clouds from~\cite{Turk:1994:ZPM}.
We randomly select a source and target subset from the original point cloud, each containing $M$ points. The target point cloud is rotated around a randomly selected axis by an angle of $\theta_{rot}$. It is then translated along a random axis to a distance of $\Delta_{trans}$. We then run the registration algorithms on the point-cloud pair and record their translational and angular error. We repeat the experiment $T$ times and report the median error for each algorithm and each cardinality $M$ of the point clouds.

For all experiments in Figure~\ref{fig:accuracy} we set $\Delta_{trans}\!=\!0.005m$ and $T\!=\!20$. The other parameters are: Bunny: $M\in\{200,300,\ldots,1000\}$,  $\theta_{rot}\!=\!8^{\circ}$. Horse: $150\leq M \leq 1000$, $\theta_{rot}\!=\!10^{\circ}$. Dragon: $M$ up to 2000, $\theta_{rot}\!=\!5^{\circ}$. \emph{BBR-F} achieves the lowest error rate, across almost all point cloud sizes. It is followed closely by \emph{BBR-N}. Only when the point density is high, Sym-ICP performs on-par with \emph{BBR-N}. This demonstrate BBR's ability to work well with very sparse point clouds. It should also be pointed out that \emph{BBR-softBD} outperforms the comparable registration method that do not use normals (as well as G-ICP). 

\subsection{\bf{Robustness}}
\label{subsec:ExperimentsRobustness}
\begin{figure}[t]
    \includegraphics[width=4.8cm]{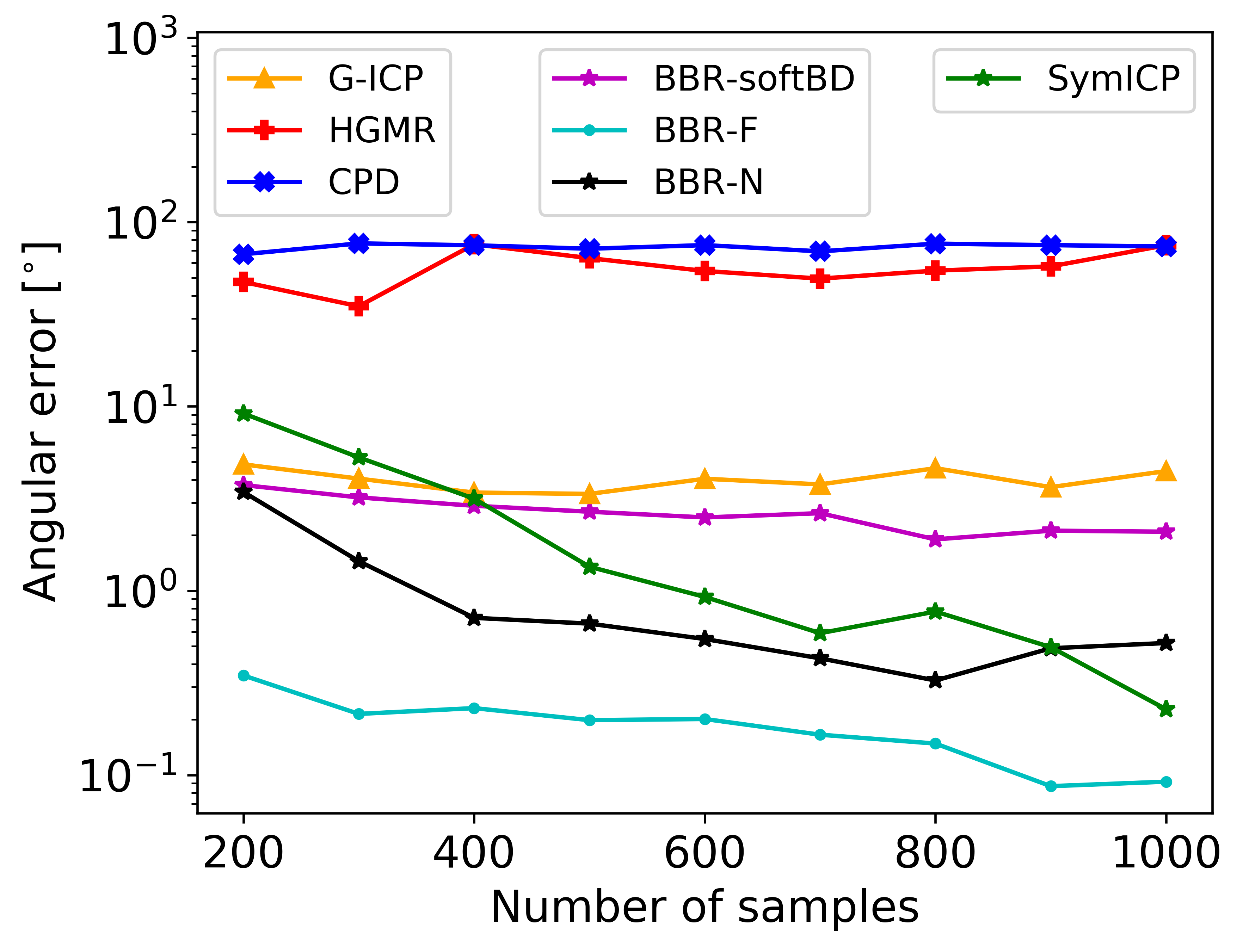}
    \includegraphics[width=5cm]{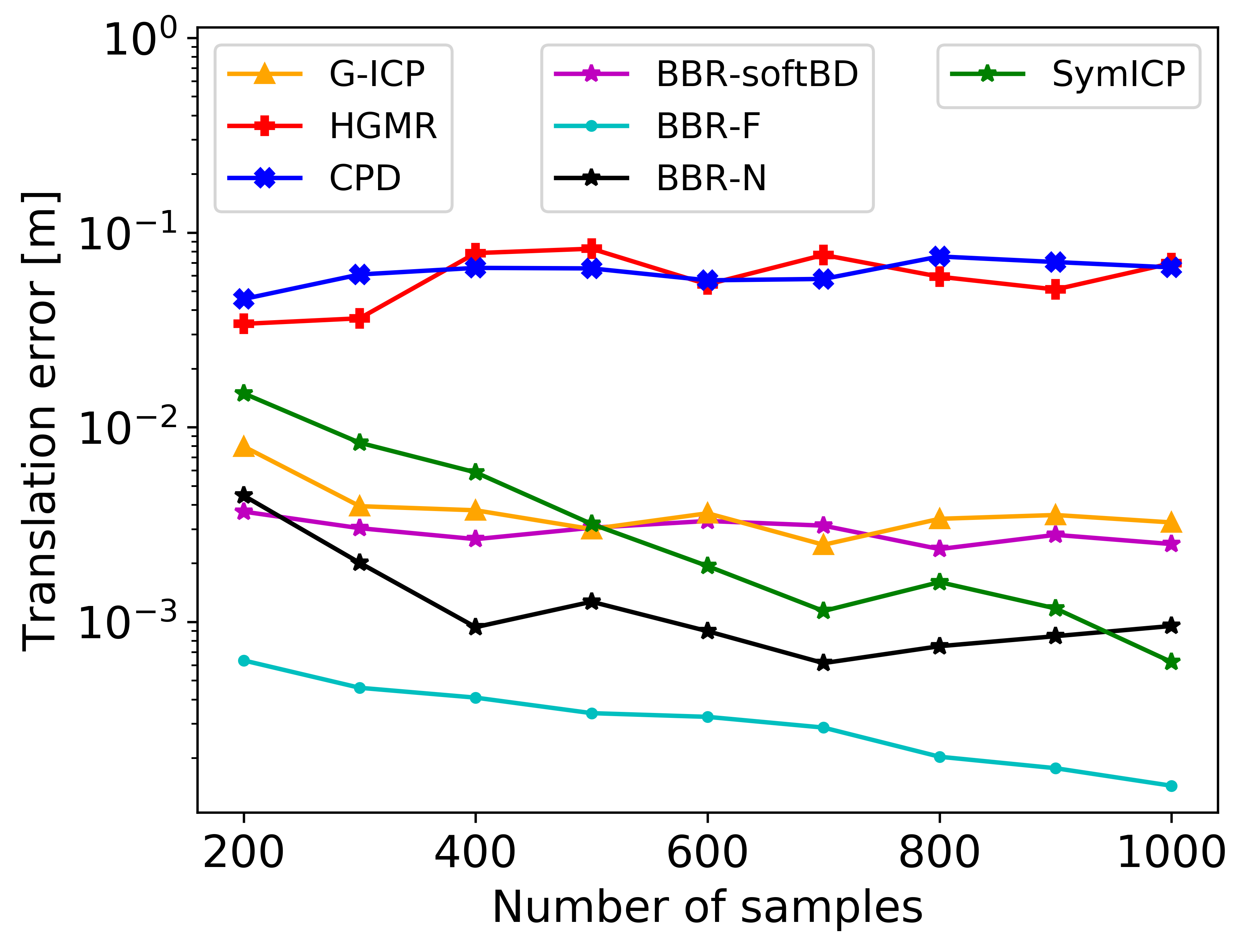}
    \includegraphics[width=2cm]{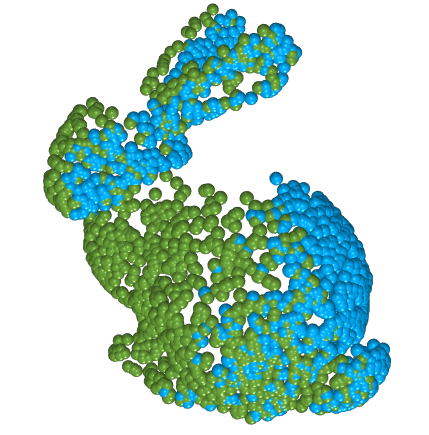} 
    \caption{{{\bf{Robustness to point of view changes:}} Right - aligned \emph{bun000} and \emph{bun090} with 1000 points, with overlap of less than 30\%. Left - angular and translation error for varying number of points. \emph{BBR-softBD}, \emph{BBR-N} and \emph{BBR-F} all outperform Sym-ICP \cite{Rusinkiewicz:2019:ASO} when the number of samples is low.}}
    \label{fig:PartialBun}
\end{figure}

The next experiments demonstrate the resistance of the algorithm to a variety of challenges, including occlusions, the presence of a distractor, measurement noise, and a large initial error.
\paragraph{{\bf Partial Overlap and Occlusion.}}
\label{subsec:RobustnessPartialBun}
The experiment shown in Figure~\ref{fig:PartialBun} evaluates the resistance to partial overlap and occlusion. We perform registration between two partial scans of the Stanford Bunny, \emph{bun000} and \emph{bun090}, each captured from a different view point. Following Rusinkiewicz's \cite{Rusinkiewicz:2019:ASO} experimental setup, we first align the scans according to the ground truth motion. Then we follow the same experimental method as in Section~\ref{subsec:Accuracy}, with $M\in\{200,300,\ldots,1000\}$, $\theta_{rot}\!=\!5^{\circ}$ and $\Delta_{trans}\!=\!0.005m$. \emph{BBR-F} achieves the most accurate results, followed by \emph{BBR-N} and \emph{Sym-ICP}. However, \emph{Sym-ICP} deteriorates considerably when given a very sparse point cloud, while both of our algorithms cope with it very well.

\begin{figure}[t]
    \centering
    \includegraphics[width=4.8cm]{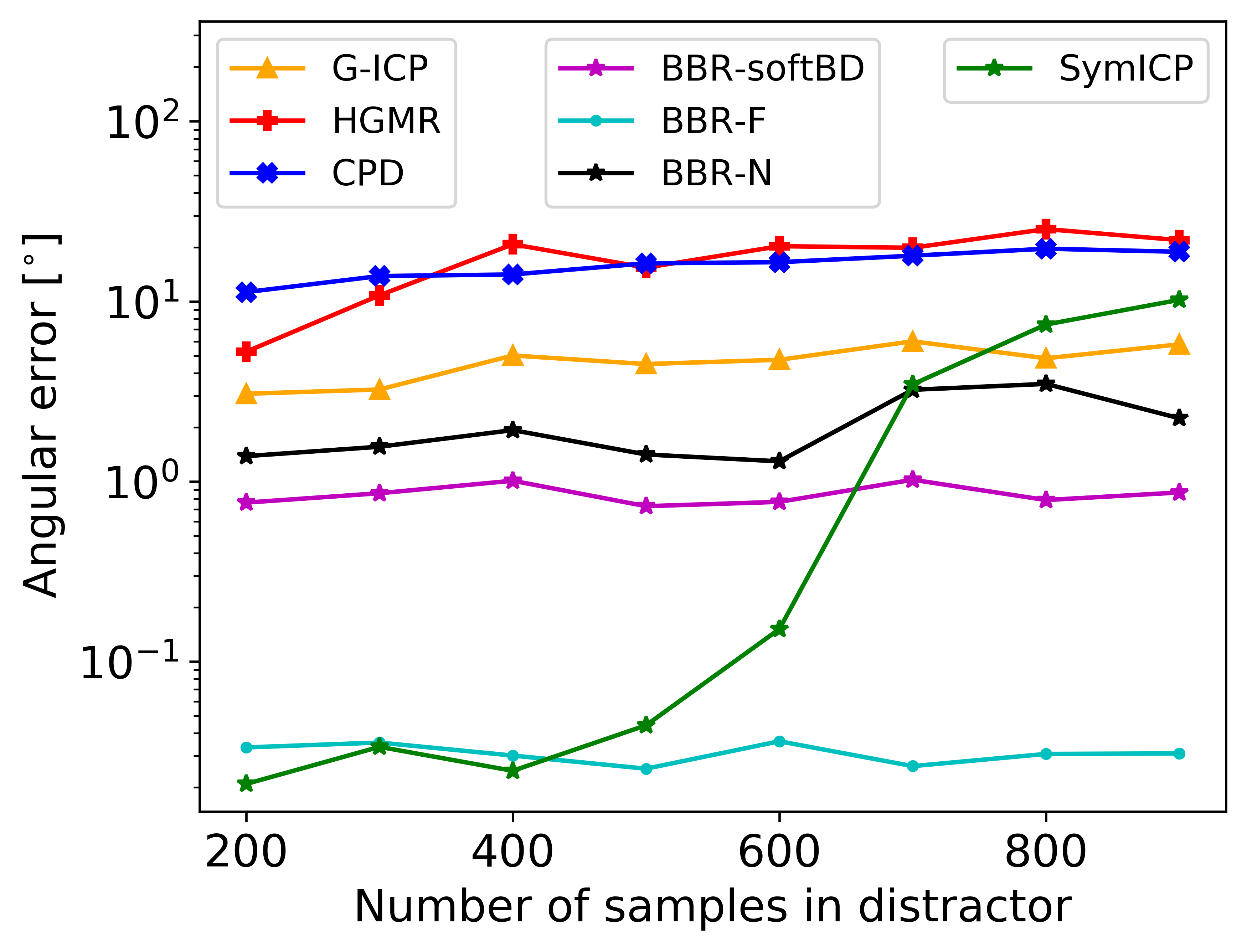}
    \includegraphics[width=5cm]{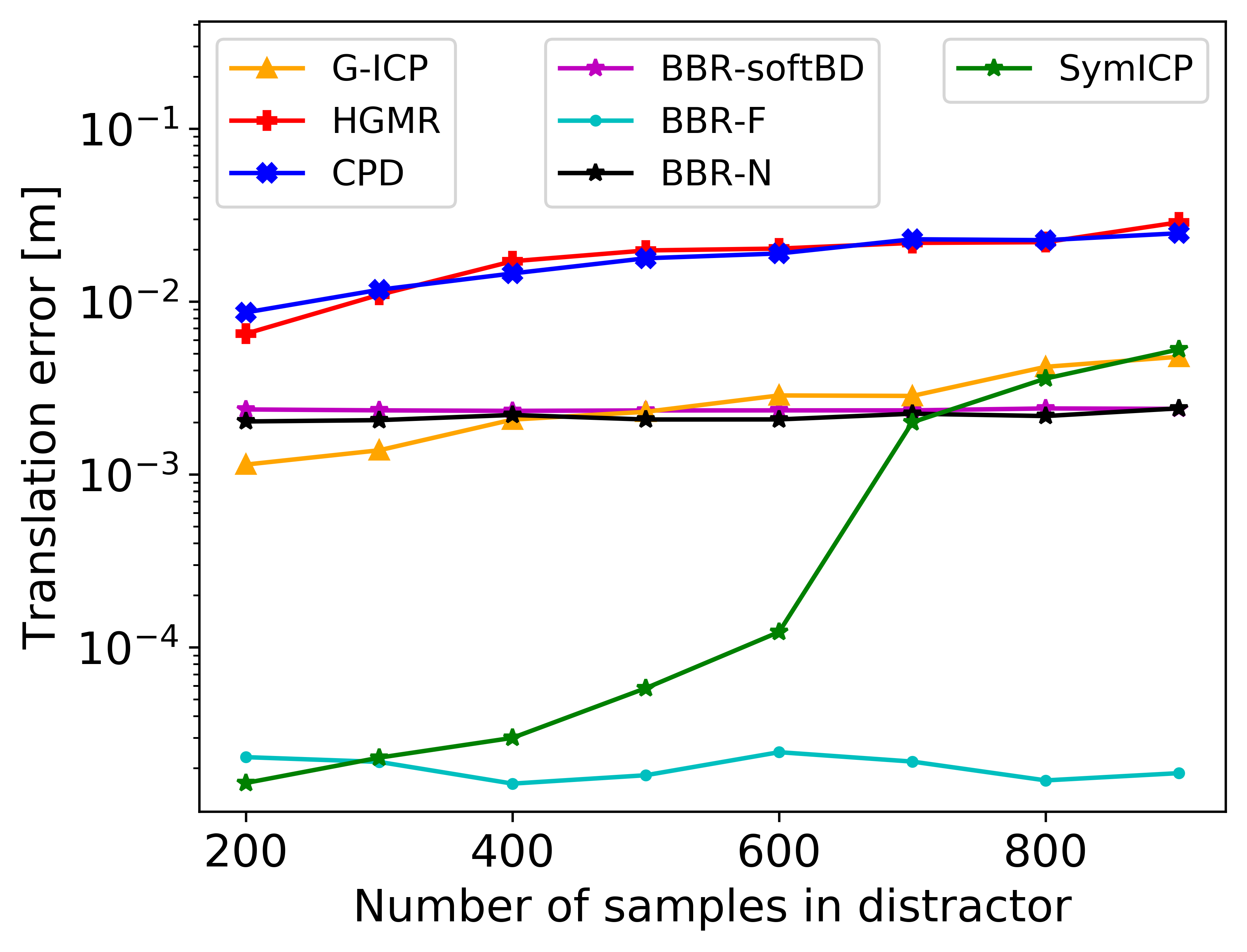}
    \includegraphics[width=2.2cm]{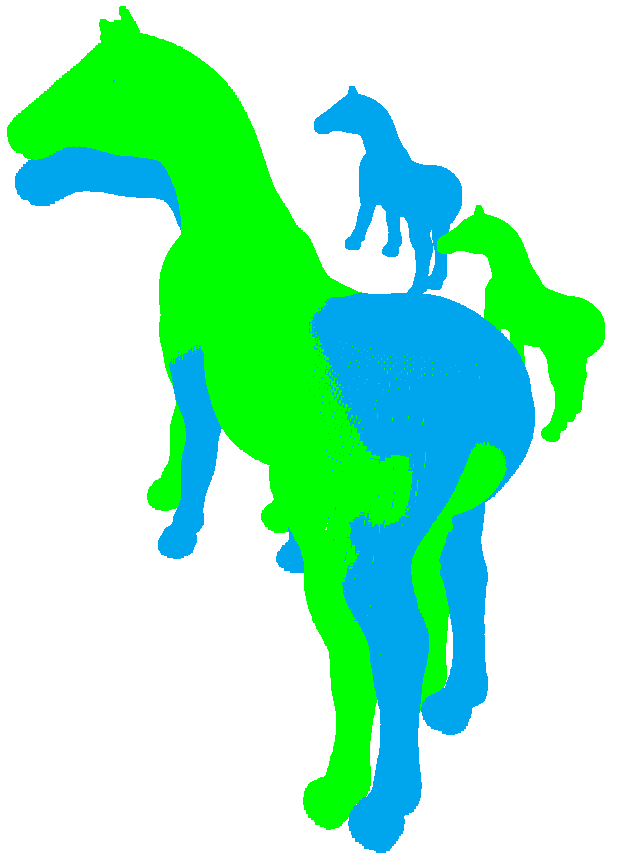}
    \caption{{\bf Distractor noise:} Right - an example of our experimental setting: The main object (large horse) has moved differently than the distractor (small horse). Left and center - the angular and translation errors for varying distractor strengths. The BBR algorithms are able to ignore the distractor and focus on the motion of the main object. }
    \label{fig:Distractor_noise}
\end{figure}

\paragraph{\bf{Resistance to Distractors.}}
This experiment evaluates the effects of distractor noise. This is the case where in addition to the main object of interest, the scene also contains a second object with a different motion. In this synthetic experiment, the main object, a large horse, was randomly translated and rotated as in Section~\ref{subsec:Accuracy}, with $T\!=\!20$, $\theta_{rot}\!=\!10^{\circ}$  and $\Delta_{trans}\!=\!0.005m$. The distractor object, a small horse, underwent a different motion. Such a situation may occur when attempting to estimate the ego-motion of a vehicle, using scans that include other independently moving vehicles. 
The main object contains 1000 points, and we vary the number of points in the distractor object from 200 up to 900. In figure~\ref{fig:Distractor_noise} we show the median error as a function of the number of points in the distractor. \emph{BBR-F} shows a strong resistance to distractor noise in this experiment, while \emph{Sym-ICP} is quite susceptible to it. Among methods that do not make use of normals, \emph{BBR-softBD} is the most accurate. 

\begin{figure}[t]
    \centering
    \includegraphics[width=5cm]{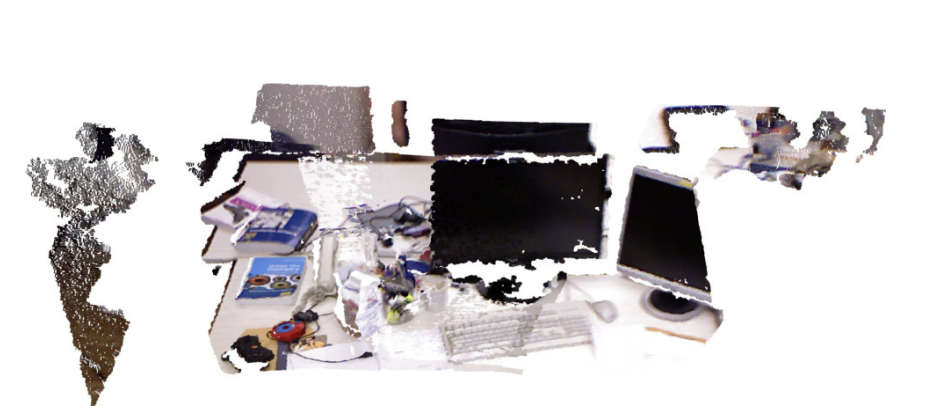}
    \includegraphics[width=3cm]{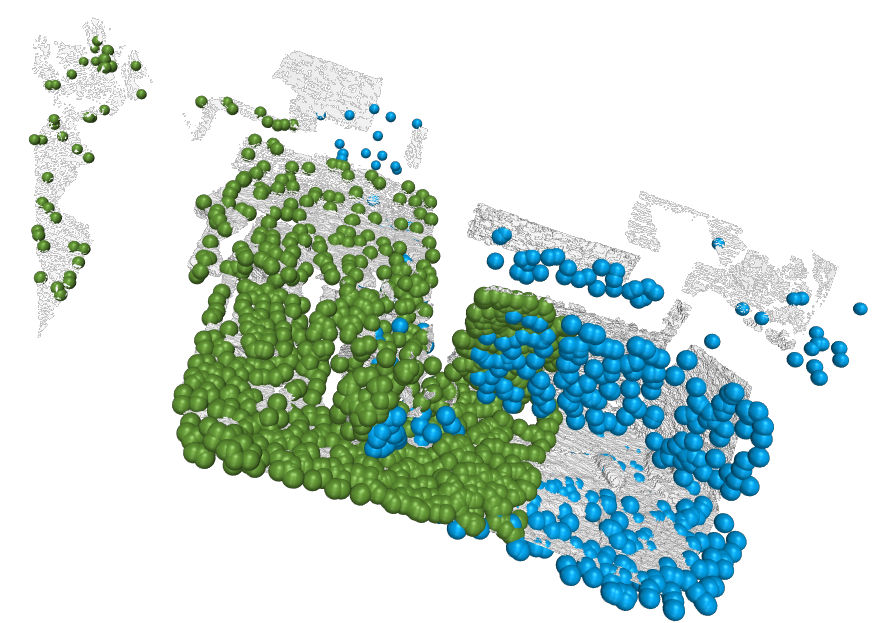}
    \begin{tabular}{cc}
    \end{tabular}
     \caption{{\bf TUM-RGB-D pair used in experiment in main paper:} Left - RGB-D image from the TUM-RGBD data set \cite{DBLP:conf/iros/SturmEEBC12}. This data is captured by the Kinect sensor, and exhibits warp and scanning noise. Right - Sampled point clouds of 1000 points of the same scene, emphasizing the small area of overlap.}
     \label{fig:TUM_images}
\end{figure}

\begin{figure}
    \centering
    \includegraphics[width=8cm]{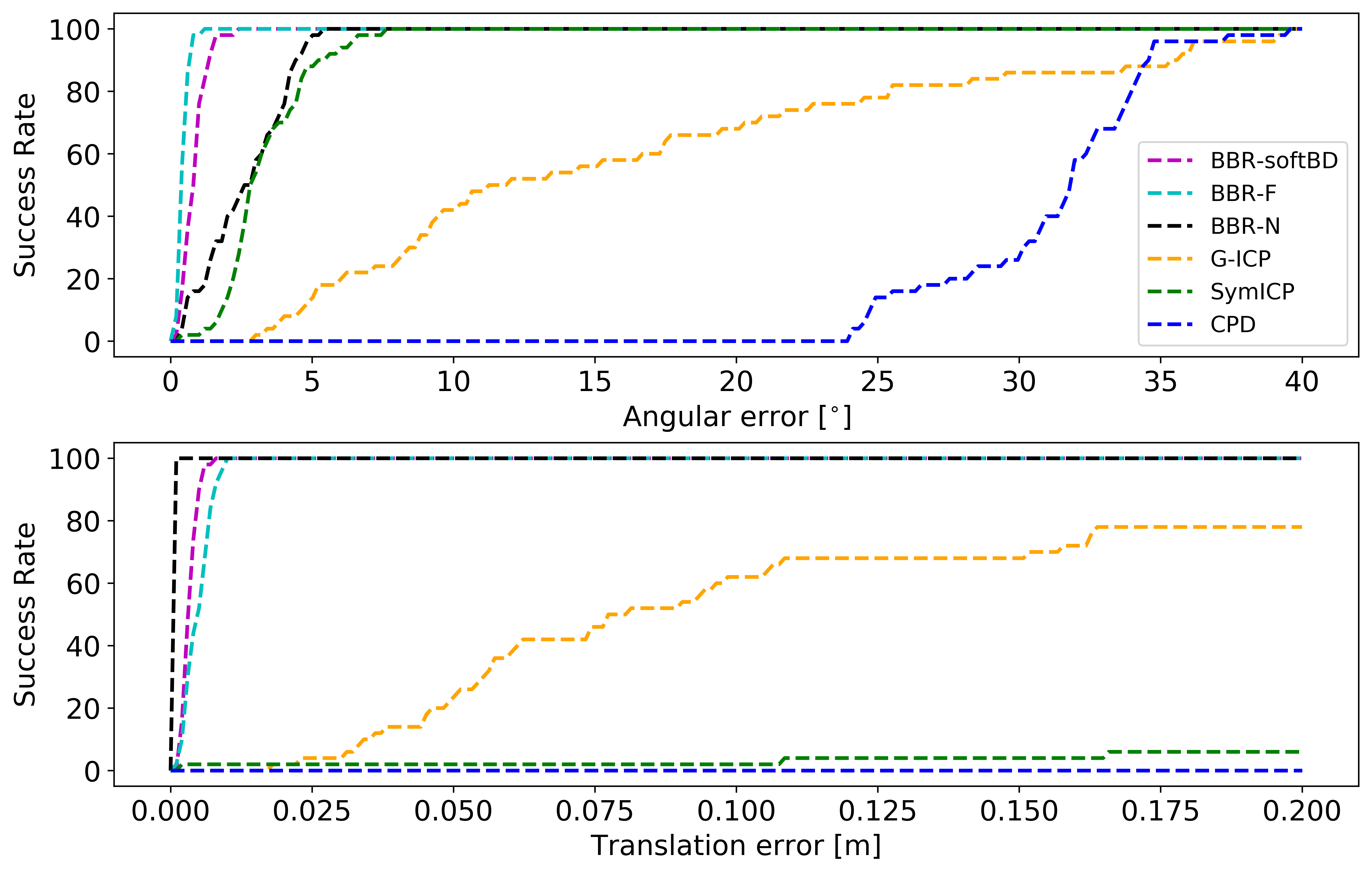}
    \caption{{\bf Convergence Analysis (TUM):} The cumulative distribution of errors over $50$ repeats. $x$-axis: the error threshold; $y$ axis: the fraction of results that achieved an error smaller than this threshold. For example, \emph{G-ICP} has $20\%$ of its registration results with an angular error lower than $5^{\circ}$, \emph{SymICP} has $90\%$ of its results under $5^{\circ}$, and the BBR algorithms have $100\%$ of their errors under $5^{\circ}$. }
    \label{fig:TUM}
\end{figure}

\paragraph{\bf{Measurement noise.}}
The TUM RGB-D dataset \cite{DBLP:conf/iros/SturmEEBC12} contains point clouds of indoor scenes captured with the Kinect sensor. It contains natural measurement noise, due to the warp and scanning noise from the Kinect sensor. It has been noted by Rusinkiewicz \cite{Rusinkiewicz:2019:ASO} that this dataset poses a qualitatively different challenge than the bunny point cloud. He demonstrates his algorithm on a specific pair of partially-overlapping scans from this dataset. We use the same pair, 1305031104.030279 and 1305031108.503548 of the {freiburg1\_xyz} sequence from the TUM RGB-D (Figure~\ref{fig:TUM_images}), sample 1000 points from each, and experiment with adding a random rotation of up to $5^{\circ}$ around a random axis. We repeat this 50 times and perform registration for each, showing the cumulative distribution of the final errors in Figure~\ref{fig:TUM}. The BBR algorithms perform better than all competing methods. 

\begin{figure}[!htb]
    \centering
    \includegraphics[width=5cm]{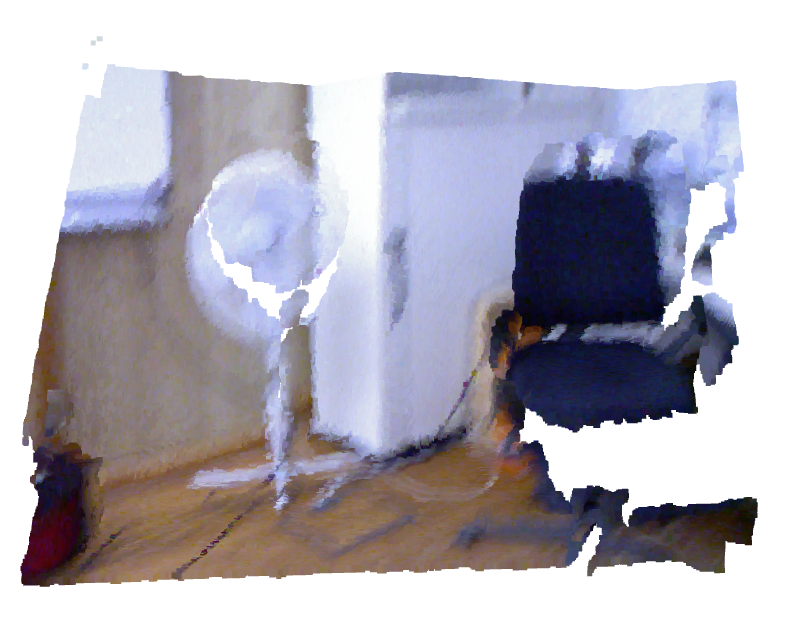}
    \includegraphics[width=5cm]{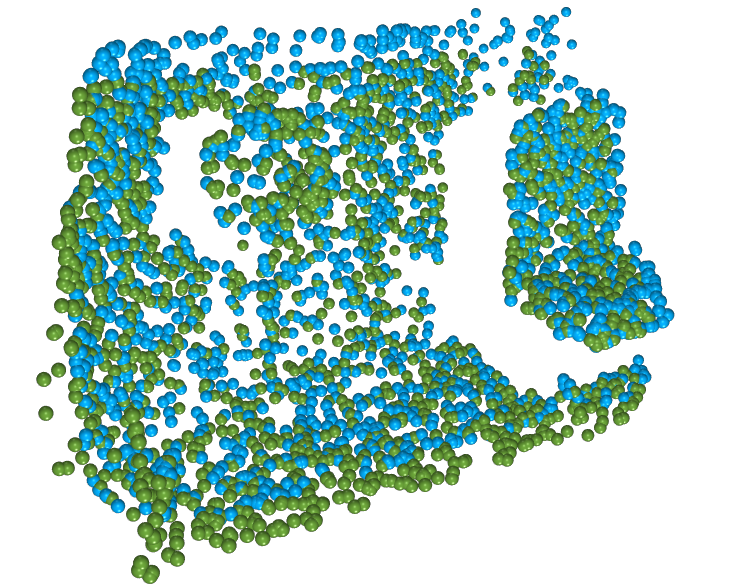}
    \begin{tabular}{cc}
    \end{tabular}
     \caption{{\textbf Additional TUM-RGB-D pair:} Left - RGB-D image from the TUM-RGBD data set \cite{DBLP:conf/iros/SturmEEBC12}. This data is captured by the Kinect sensor, and exhibits warp and scanning noise. Right - Sampled point clouds of 1500 points of the same scene.}
     \label{fig:TUM_images_2}
\end{figure}

\begin{figure}[!htp]
    \centering
    \includegraphics[width=8cm]{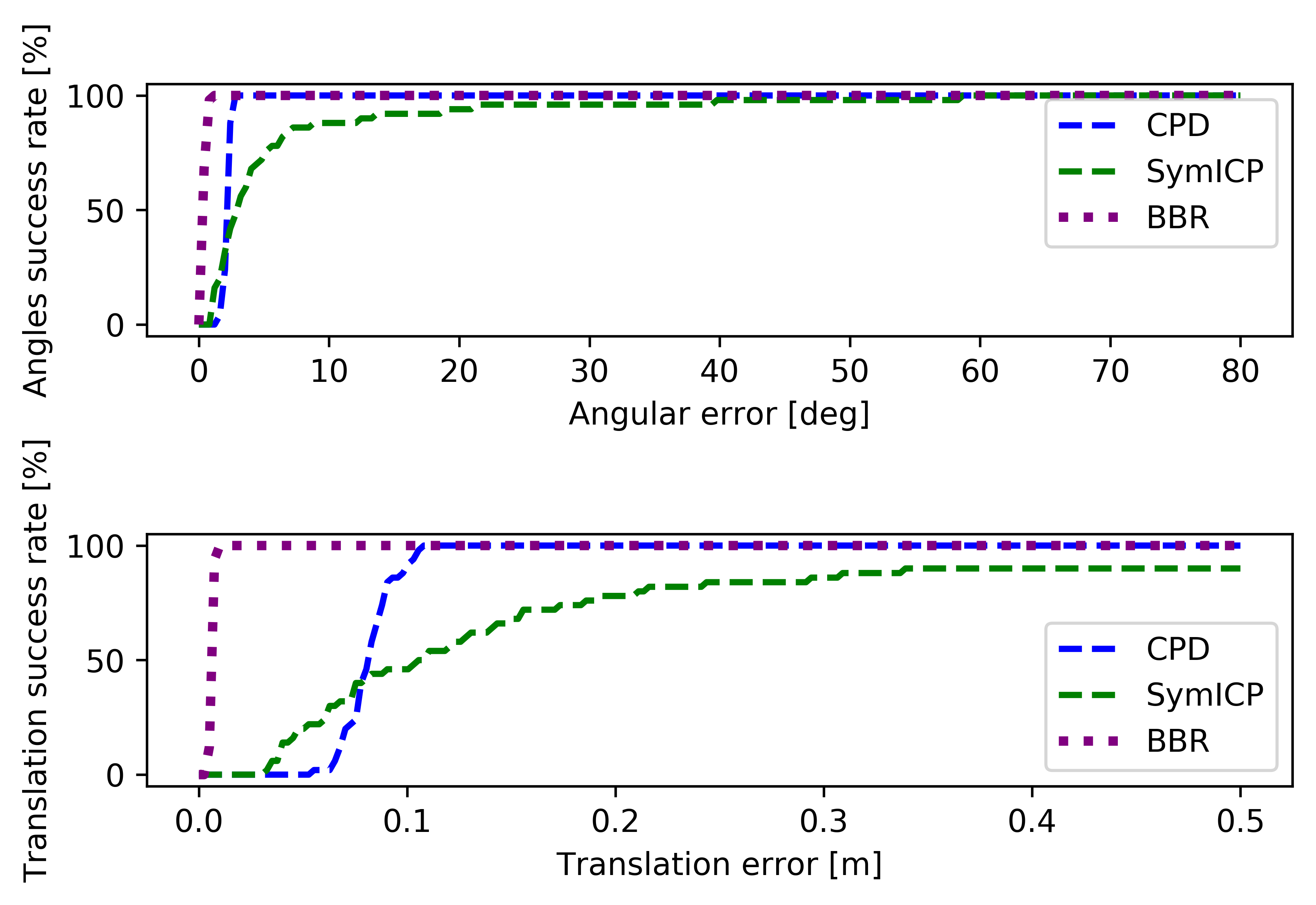}
    \caption{{\textbf Convergence Analysis (TUM):} The cumulative distribution of errors over $50$ repeats. The $x$-axis is the error threshold and the $y$ axis is the fraction of results that achieved an error smaller than this threshold. \emph{BBR-softBD} is labeled \textbf{BBR}}
    \label{fig:TUM_2}
\end{figure}

We test another pair of scans from TUM RGB-D, 1305031794.813436 and 1305031794.849048 of the {freiburg1\_xyz} sequence (Figure ~\ref{fig:TUM_images_2}). We sample 1500 points from each, and experiment with adding a random rotation of up to $5^{\circ}$ around a random axis. We repeat this 50 times and perform registration for each, showing the cumulative distribution of the final errors in Figure~\ref{fig:TUM_2}. \emph{BBR-softBD} (labeled \textbf{BBR}) performs considerably better than either Sym-ICP~\cite{Rusinkiewicz:2019:ASO} or CPD~\cite{Myronenko:2010:PSR}, showing its robustness to realistic measurement noise and occlusions. 

\begin{figure}[t]
    \centering
    \includegraphics[width=8cm]{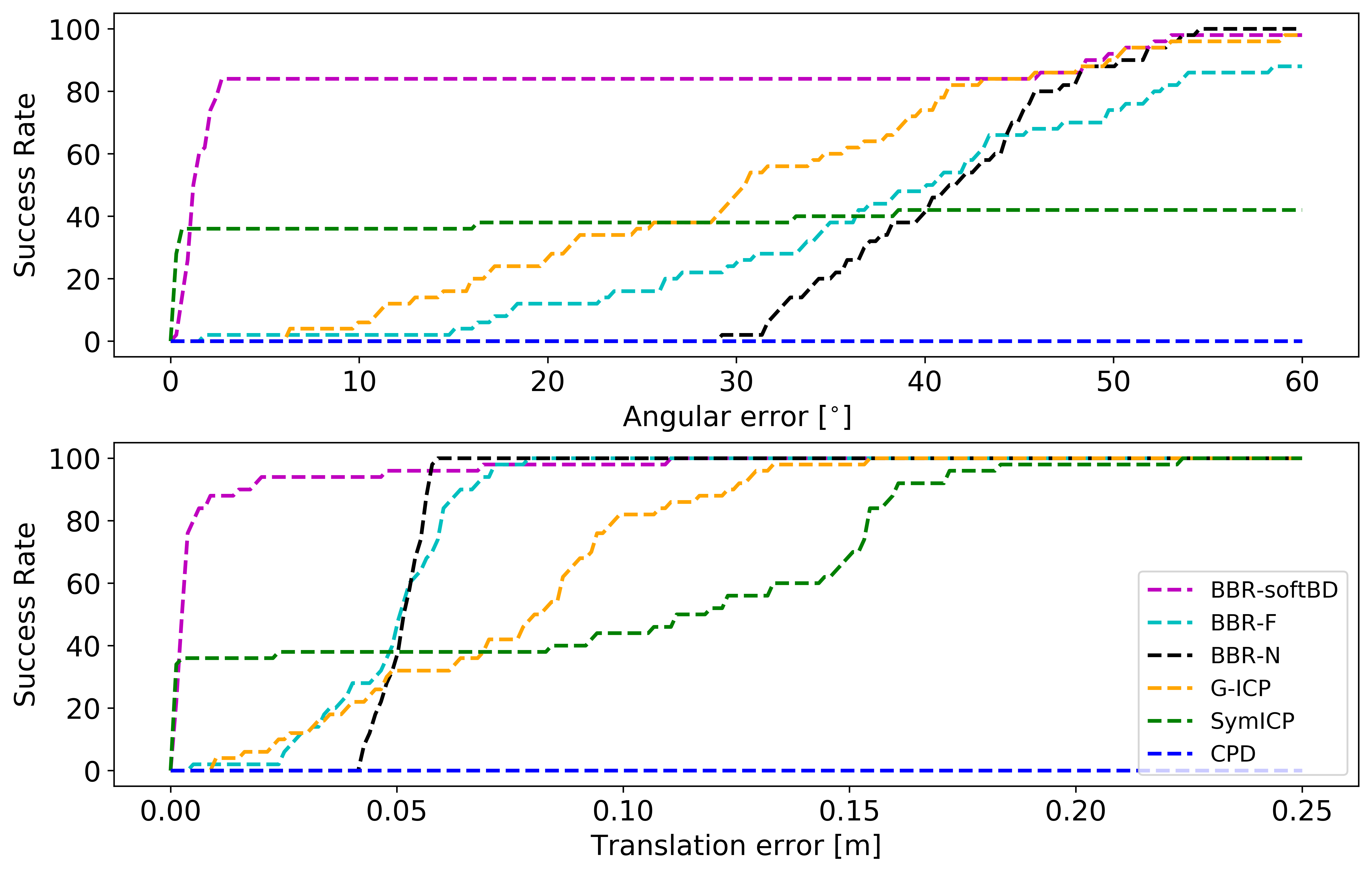}
    \includegraphics[width=3cm]{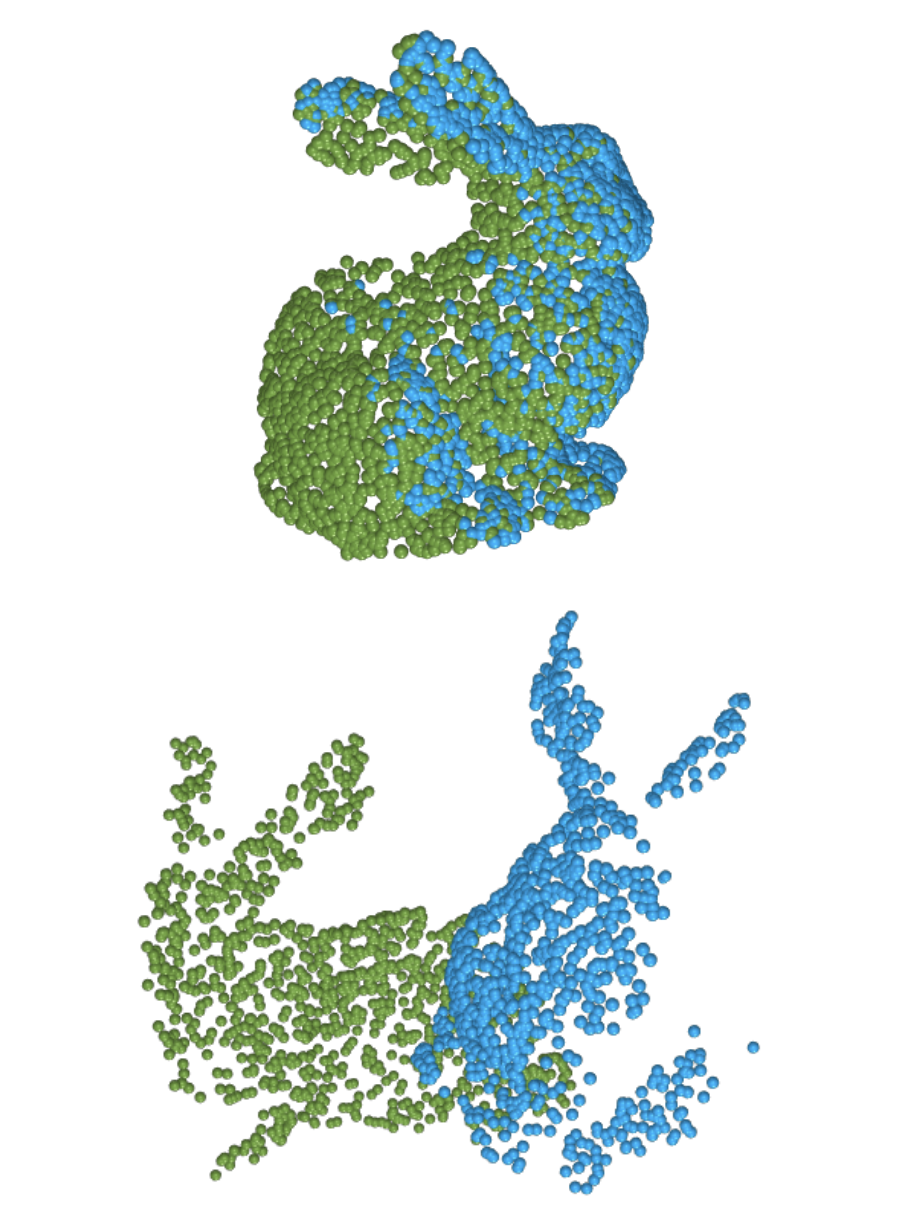}
    \caption{{\bf Convergence Analysis (partially-overlapping Bunny):} Right: \emph{bun180} and \emph{bun270}, aligned (top), and in their initial position: rotated in $50^{\circ}$ and translated by a factor of $0.05m$ (bottom). Left: the cumulative distribution of errors over $50$ repeats. The $x$-axis is the error threshold and the $y$ axis is fraction of registration results whose error is smaller than this threshold. }
    \begin{tabular}{cc}
    \end{tabular}
    \label{fig:ConvergencePartialBun}
\end{figure}

\paragraph{\bf Large initial error with partial overlap.}
In this experiment we evaluate the ability to converge when the source and target point clouds are only partially overlapping, and starting from a large initial error. We use two scans of the bunny point cloud, \emph{bun180} and \emph{bun270}, that were captured from significantly different view points. As before, we first align them, sample 1000 points from each, and then apply a random motion to one of them. In this experiment the motion is large: a random rotation in the range $[30,50]$ degrees, around a random axis, and a translation of half the extent of the point cloud ($0.05m$) along a random direction. Figure~\ref{fig:ConvergencePartialBun} shows the cumulative distribution of errors over 50 repeats. \emph{BBR-softBD} performs considerably better than all other algorithms. 

\begin{figure}[!htp]
    \centering
    \hspace*{0.5cm}\includegraphics[width=2.5cm]{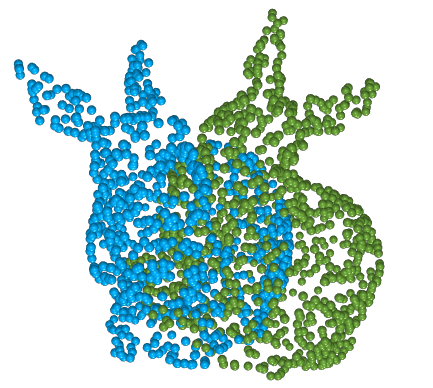}
    \hspace*{1cm}\includegraphics[width=3cm]{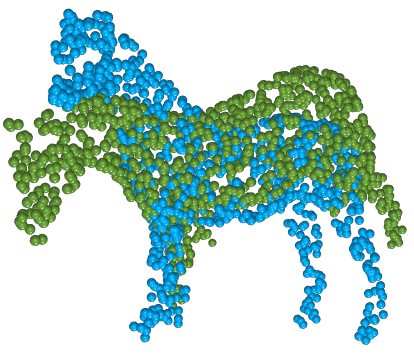}
    \hspace*{1cm}\includegraphics[width=3cm]{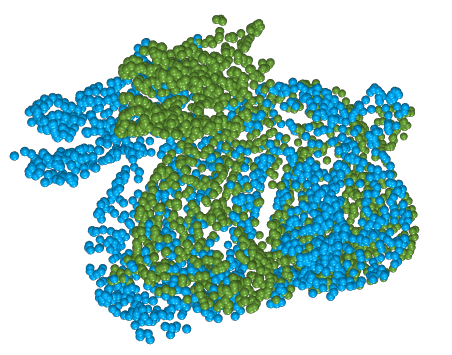}\\
    \includegraphics[width=3.9cm]{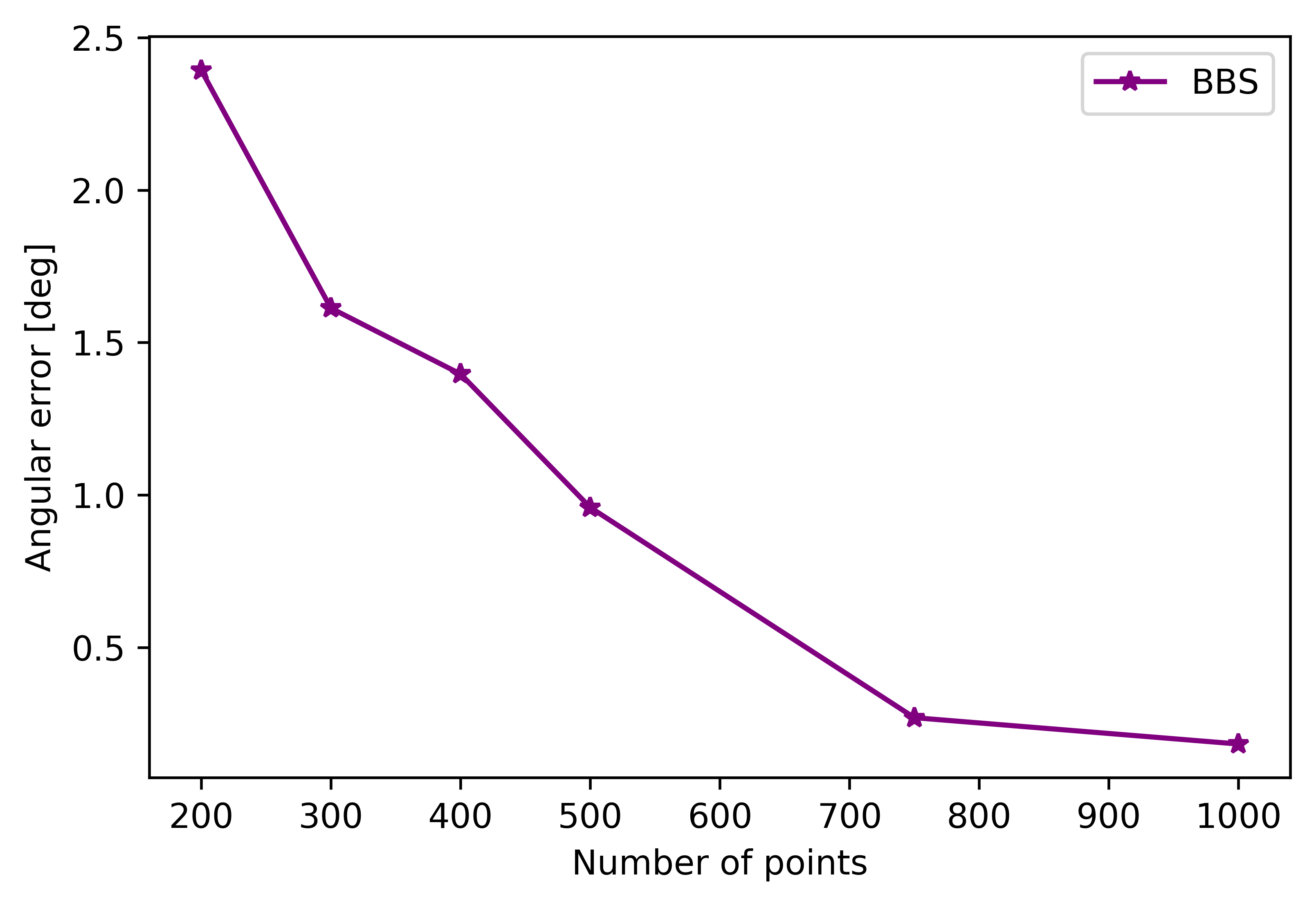} 
    \includegraphics[width=4cm]{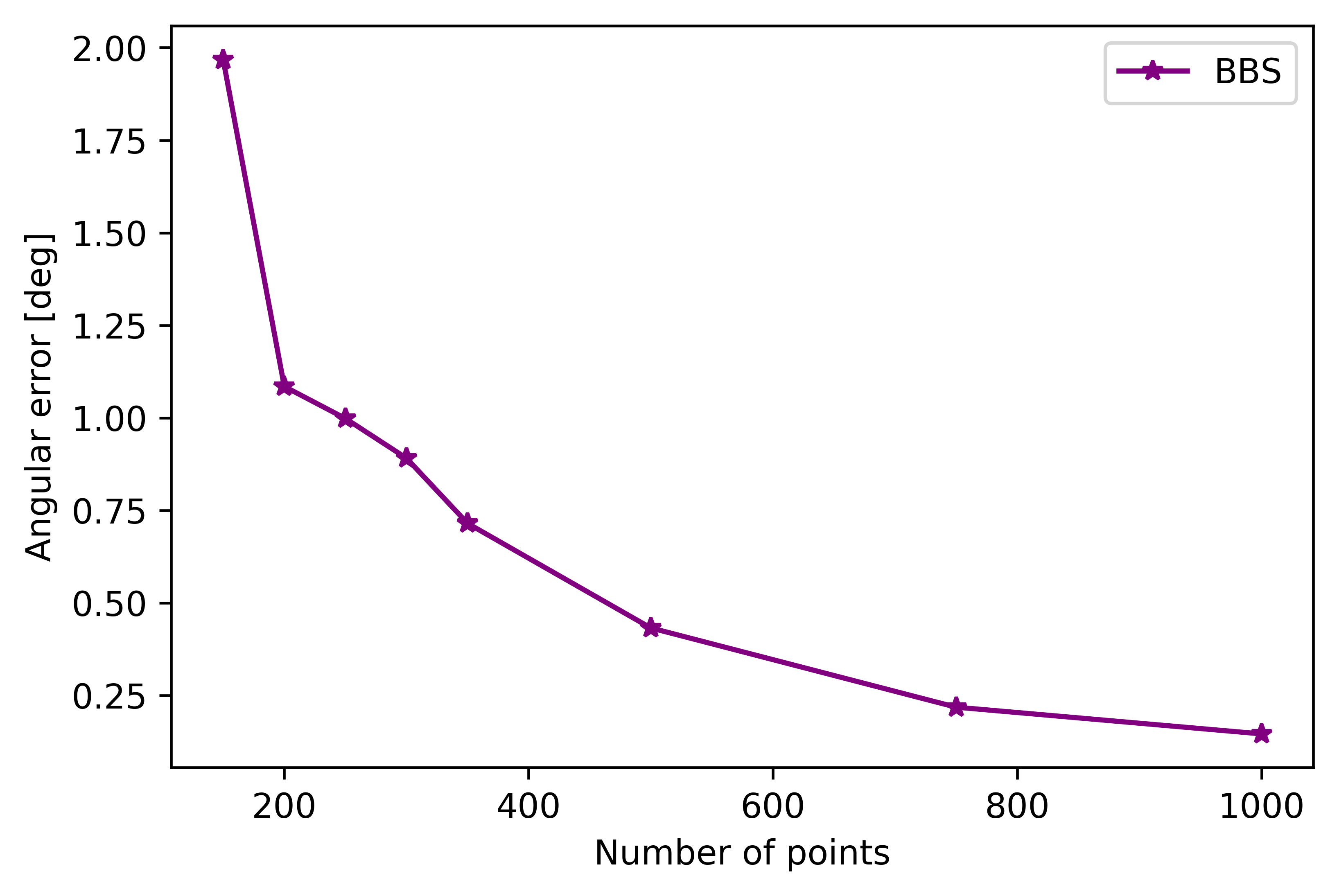}
    \includegraphics[width=4.05cm]{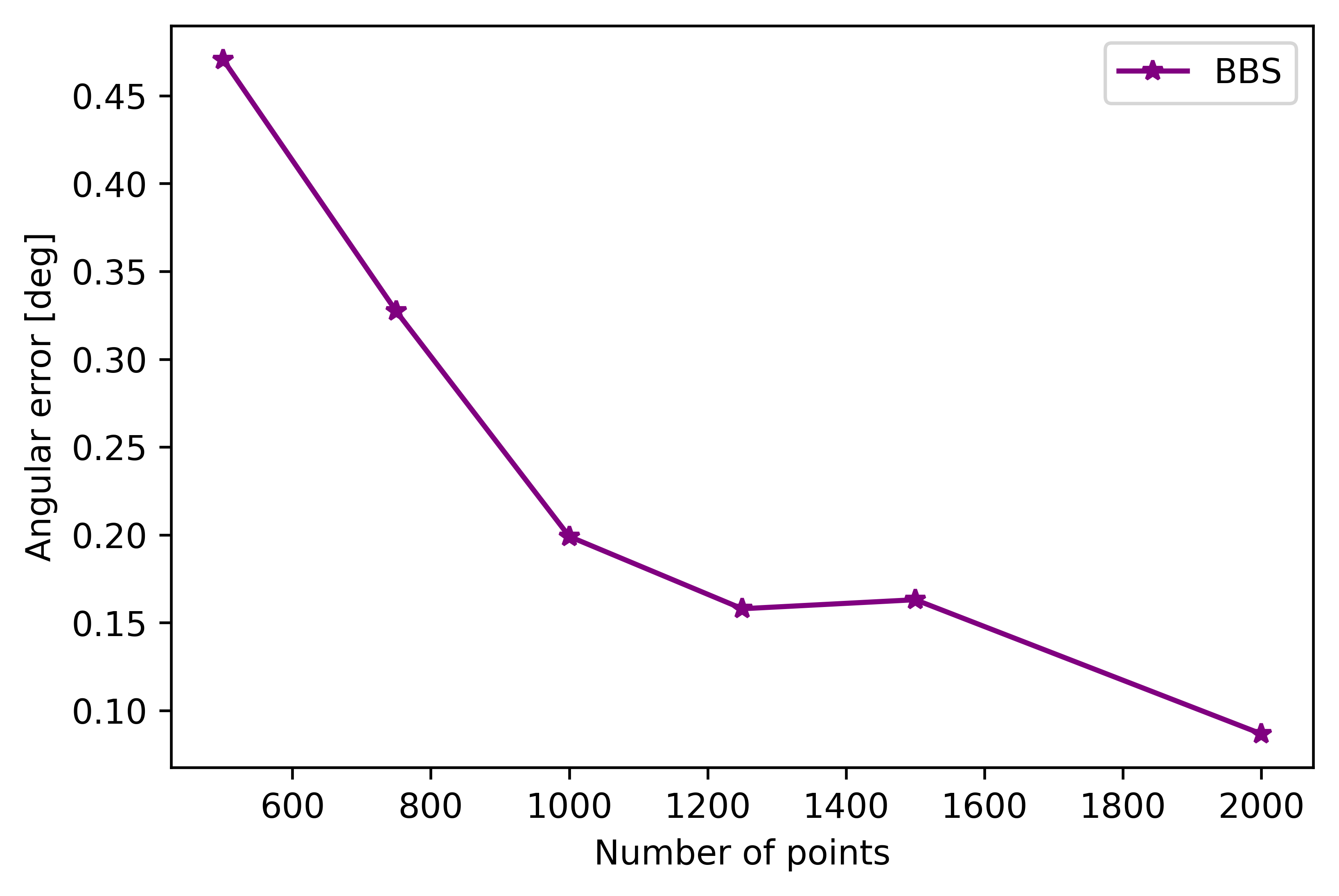} \\
    \includegraphics[width=3.9cm]{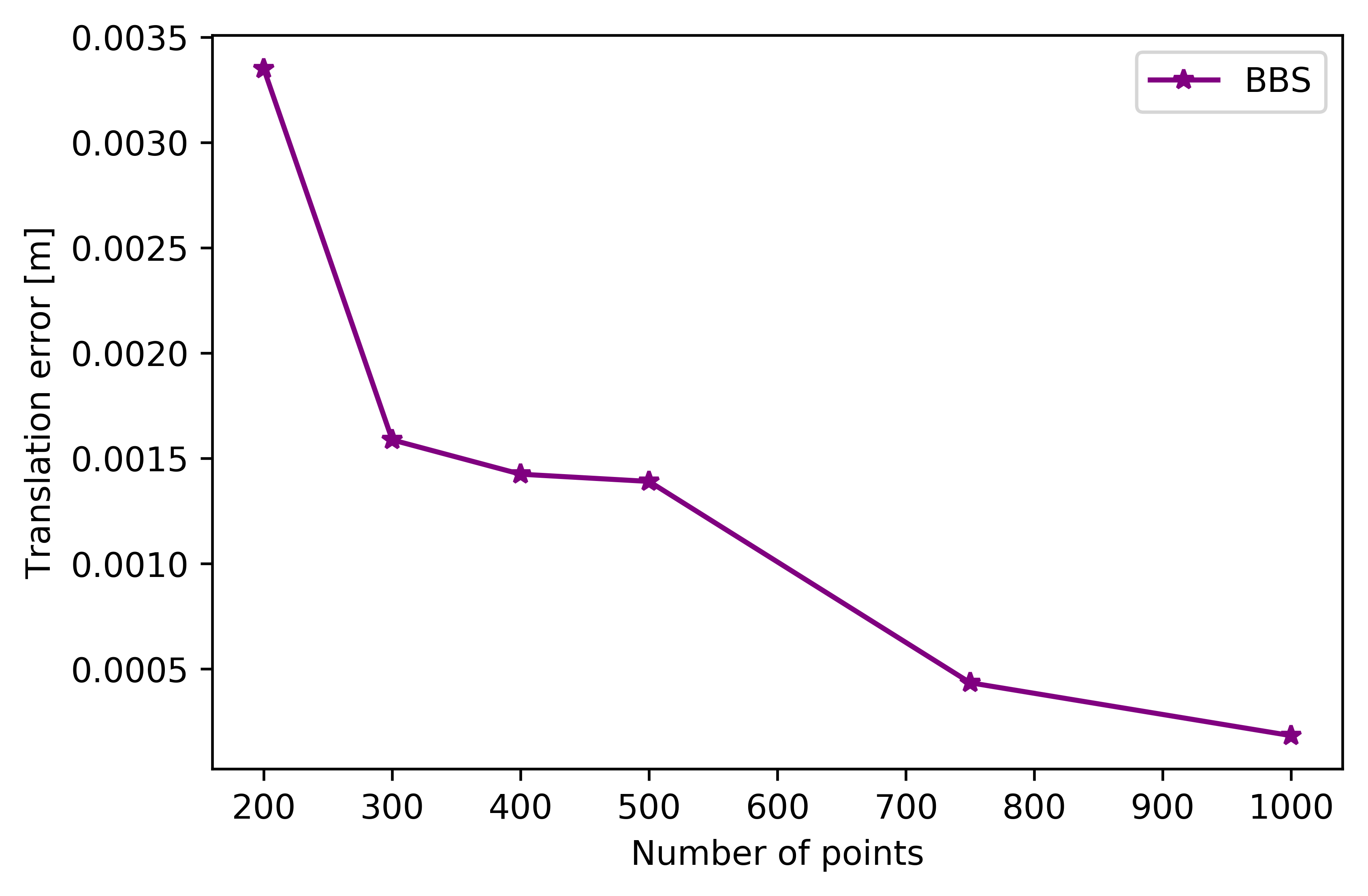} 
    \includegraphics[width=4cm]{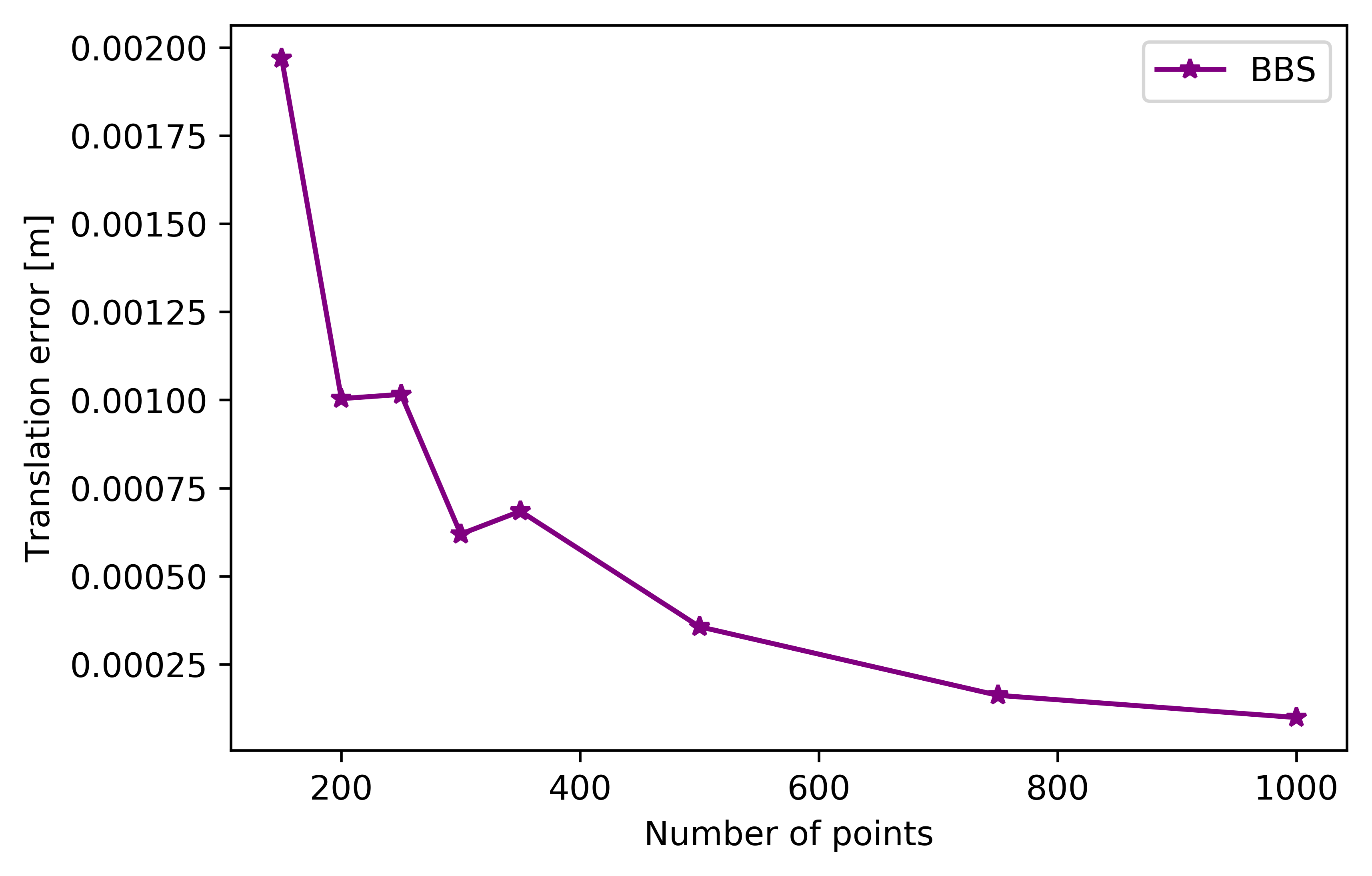}
    \includegraphics[width=4.05cm]{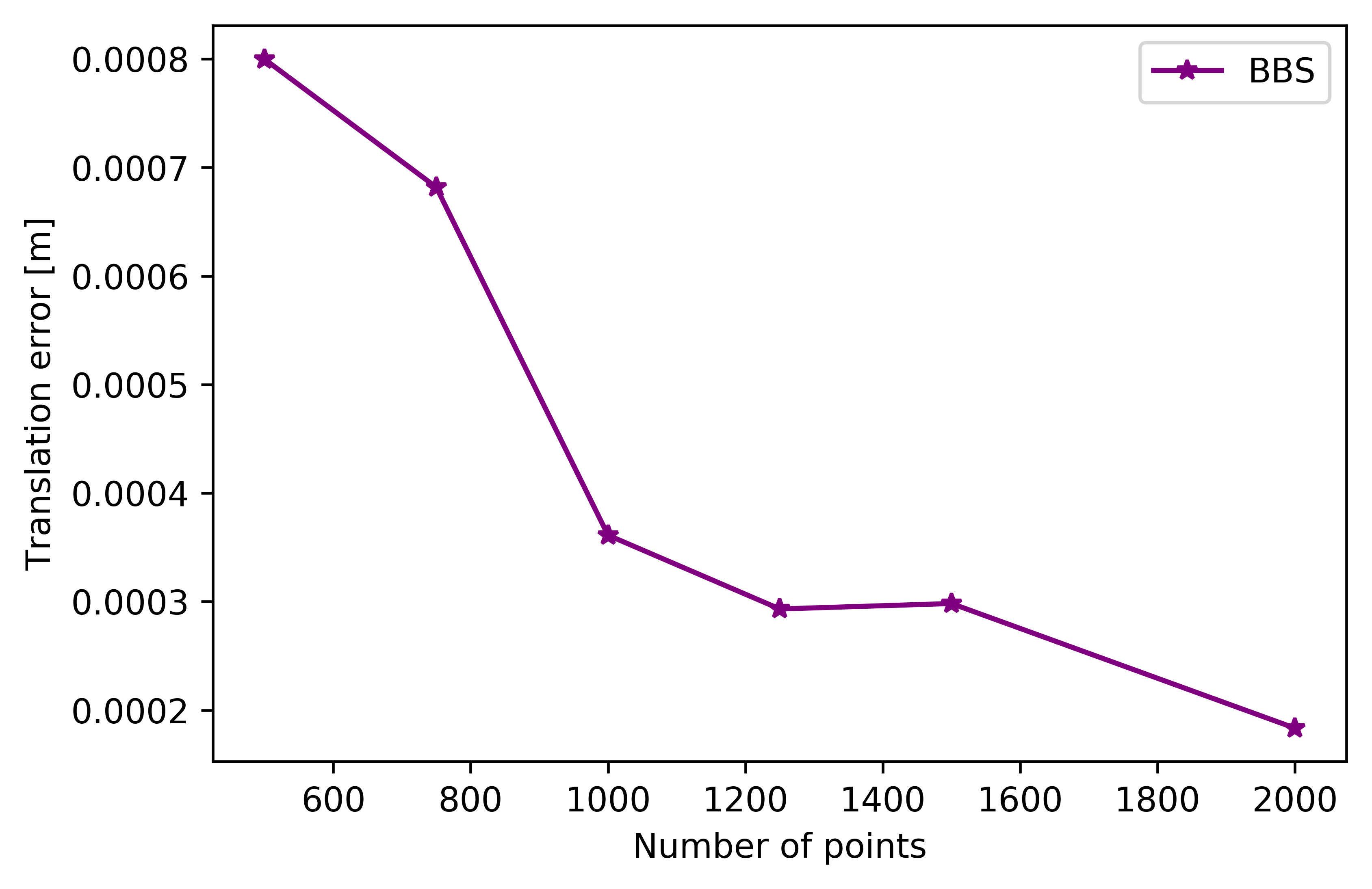}
     \caption{{\textbf Convergence test:} Top - Point clouds examples as used in the accuracy test. In this experiment, $\theta_{rot}\!$ was randomized in range $=[30, 60]$ degrees. In the point cloud visualizations above, the Bunny (left) is rotated by $50^{\circ}$ . The Horse (middle) is rotated by $30^{\circ}$. The Dragon (right) is rotated by $45^{\circ}$. Center and Bottom - angular and translation error as a function of the number of points. In all cases, \emph{BBR-softBBS} (labeled \textbf{BBS}) manages to reduce the large initial rotation error significantly.}
    \label{fig:convergence}
\end{figure}

\paragraph{\textbf{BBR-softBBS Convergence.}}
We've previously demonstrated the ability of \emph{BBR-softBD}, \emph{BBR-N} and \emph{BBR-F} to handle cases where the initial error is of the order of up to $10^{\circ}$. Here we show that \emph{BBR-softBBS} can be used to register in situations where the initial error is much larger. This is due to its large basin of convergence. In this section we use the same Bunny, Horse and Dragon models that were used in the accuracy test, this time with a large random rotation in range $\theta_{rot}\!\in[30, 60]$ degrees, and run \emph{BBR-softBBS}. We set $\Delta_{trans}\!=\!0.005m$ and $T\!=\!20$. As can be seen in Figure~\ref{fig:convergence}, \emph{BBR-softBBS} manages to reduce the rotation error significantly, to below $3^{\circ}$, in all experiments. 

\subsection{Odometry}
\label{subsec:Odometry}

In this section we present experiments in the realistic setting of vehicle navigation, specifically focusing on a difficult setting where the separation between the two clouds is relatively large, leading to significant occlusions and outliers caused by independently moving objects in the scene. We use two datasets of high-resolution Lidar scans: KITTI Odometry~\cite{Geiger2012CVPR}, and Apollo-Southbay~\cite{apollo}, both of which  consist of large point clouds of over 100K points. We follow the experimental setup of~\cite{Lu_2019_ICCV}, where the test set consists of pairs of clouds scanned by the same vehicle at two times, during which the vehicle has travelled up to 5 meters, for up to 2 seconds (KITTI) or even 5 seconds (Apollo-Southbay). An initial estimate for the motion is assumed to be available, which is inaccurate by up to 1 meter in each of x,y,z, and up to 1 degree in each of $\theta, \phi, \psi$. We report the mean translation and angular errors, as well as the maximum (worst case). 

For this test, we use the \emph{BBR-F} variant of our BBR algorithm. Normals are estimated from neighborhoods of $k=94$ points in the full cloud. To achieve high accuracy, we only moderately subsample the point clouds to 30K points, uniformly at random. 

Tables~\ref{tab:kitti} and ~\ref{tab:apollo} compare our results to those reported in~\cite{Lu_2019_ICCV}, and to those of Sym-ICP~\cite{Rusinkiewicz:2019:ASO}. In both experiments, \emph{BBR-F} achieves state of the art accuracy by three of the four accuracy measures. This demonstrates BBR's capabilities of achieving high accuracy in realistic scenarios that include occlusions, outlier motions and measurement noise.

\begin{table}
	\small
	\begin{center}
		\begin{tabular}{|p{4.0cm}|cc|cc|}
			\toprule[1pt]
			\multirow{2}{*}{\makecell[tl]{Method}} & Angular & Error($^\circ$) & Translation & Error(m) \\
			\cline{2-3}\cline{4-5}
			& Mean & Max & Mean & Max  \\
			\midrule[.5pt]
			ICP-Po2Po \cite{ICP}                   & 0.139 & 1.176  & 0.089  & 2.017   \\
			ICP-Po2Pl \cite{ICP}                   & 0.084 & 1.693  & 0.065  & 2.050   \\
			G-ICP \cite{SegalHT09}               & 0.067 & 0.375  & 0.065  & 2.045   \\
			AA-ICP \cite{Pavlov2018}             & 0.145 & 1.406  & 0.088  & 2.020   \\
			NDT-P2D \cite{journals/ijrr/Stoyanov0AL12}             & 0.101 & 4.369  & 0.071  & 2.000   \\
			CPD \cite{Myronenko:2010:PSR}             & 0.461 & 5.076  & 0.804  & 7.301   \\
			3DFeat-Net \cite{yew20183dfeat}            & 0.199 & 2.428  & 0.116  & 4.972   \\
			DeepVCP-Base \cite{Lu_2019_ICCV}                               & 0.195  & 1.700  & 0.073   & \textbf{0.482}   \\
			DeepVCP-Duplication  \cite{Lu_2019_ICCV}                                & 0.164  & 1.212  & 0.071  & \textbf{0.482}   \\
			Sym-ICP \cite{Rusinkiewicz:2019:ASO} & 0.066  & 0.422  & \textbf{0.058}   & 0.863   \\
			\midrule[.5pt]
			BBR-F (ours)         & \textbf{0.065}  & \textbf{0.356}  & \textbf{0.058}   & 0.730   \\
			\bottomrule[1pt] 
		\end{tabular}
	\end{center}
	\caption{
		The KITTI Odometry dataset. Our algorithm achieves state of the art results for almost all accuracy measures. 
	}
	\label{tab:kitti}
\end{table}

\begin{table}
	\small
	\begin{center}
		\begin{tabular}{|p{4.0cm}|cc|cc|}
			\toprule[1pt]
			\multirow{2}{*}{\makecell[tl]{Method}} & Angular & Error($^\circ$) & Translation & Error(m) \\
			\cline{2-3}\cline{4-5}
			& Mean & Max & Mean & Max  \\
			\midrule[.5pt]
			ICP-Po2Po \cite{ICP}                   & 0.051 & 0.678  & 0.089  & 3.298   \\
			ICP-Po2Pl \cite{ICP}                   & 0.026 & 0.543  & 0.024  & 4.448   \\
			G-ICP \cite{SegalHT09}               & 0.025 & 0.562  & 0.014  & 1.540   \\
			AA-ICP \cite{Pavlov2018}             & 0.054 & 1.087  & 0.109  & 5.243   \\
			NDT-P2D \cite{journals/ijrr/Stoyanov0AL12}             & 0.045 & 1.762  & 0.045  & 1.778   \\
			CPD \cite{Myronenko:2010:PSR}             & 0.054 & 1.177  & 0.210  & 5.578   \\
			3DFeat-Net \cite{yew20183dfeat}             & 0.076 & 1.180  & 0.061  & 6.492   \\
			DeepVCP-Base \cite{Lu_2019_ICCV}                                & 0.135  & 1.882  & 0.024   & \textbf{0.875}   \\
			DeepVCP-Duplication \cite{Lu_2019_ICCV}                                & 0.056  & 0.875  & 0.018   & 0.932  \\
            Sym-ICP \cite{Rusinkiewicz:2019:ASO} & 0.018  & 2.589  & 0.010   & 6.625   \\
			\midrule[.5pt]			
			BBR-F (ours)                                & \textbf{0.015}  & \textbf{0.308}  & \textbf{0.007}   & 2.517  \\			
			\bottomrule[1pt] 
		\end{tabular}
	\end{center}
	\caption{
		The Apollo-SouthBay odometry dataset. Our algorithm achieves state of the art results for almost all accuracy measures.  
	}
	\label{tab:apollo}
\end{table}

\subsection{Run-time}
\label{subsec:runtime}

Table~\ref{tab:BBR_time} shows the time it takes to run a single iteration of the different variants of our loss function, as a function of the number of points (measured with a PyTorch implementation running on a GTX 980 Ti GPU). All of the algorithm variants typically converge in few hundreds of iterations, depending on the learning rate. 
 
\begin{table}
\begin{center}
\begin{tabular}{|c|c|c|c|c|}
\hline
Points & \emph{BBR-softBBS} & \emph{BBR-softBD} & \emph{BBR-N} & \emph{BBR-F}  \\
\hline
200         &                  4     & 4     &       4  & 6        \\
500         &                  4     &  4     &       6     & 6    \\
1000         &                 8     &  8     &      13      & 8   \\
5000        &                 140     & 140     &       240  & 24          \\
30000        &                 -     & -     &       -  & 125          \\
\hline
\end{tabular}
\caption{Average running time of a single gradient descent iteration in milliseconds, as a function of the number of points in the cloud. Due to memory limitations, only BBR-F is able to handle 30000 points.}
\label{tab:BBR_time}
\end{center}
\end{table}

\section{Conclusions}
We proposed \emph{Best Buddy Registration (BBR)} algorithms for point cloud registration inspired by the Best Buddy Similarity (BBS) measure. First we show that registration can be performed by running gradient descent on a differential approximation to the negated BBS measure. This results in an algorithm that is quite robust to noise, occlusions and distractions, and able to cope with very sparse point clouds. We then present additional algorithms that achieve higher accuracy while maintaining robustness, by incorporating point-to-point and point-to-plane distances into the loss function. Finally, we present the \emph{BBR-F} algorithm that uses \emph{best buddy filtering} to achieves state of the art accuracy on challenges that include significant noise, occlusions and distractors, including registration of automotive lidar scans that are relatively widely separated in time. Our algorithms are implemented in Pytorch and optimized with Adam gradient descent, allowing them to be incorporated as a registration stage in Deep Neural Networks for processing point clouds.

\section{Acknowledgements}
This research was supported by ERC-StG grant no. 757497 (SPADE) and by ISF grant number 1549/19. 

\bibliographystyle{splncs}
\bibliography{BBRbib}
\end{document}